\documentclass[lettersize,journal]{IEEEtran}
\usepackage{amsmath,amsfonts}
\usepackage{booktabs}
\usepackage{multirow}
\usepackage{algorithmic}
\usepackage{algorithm}
\usepackage{array}
\usepackage[printonlyused]{acronym}
\usepackage[caption=false,font=normalsize,labelfont=sf,textfont=sf]{subfig}
\usepackage{textcomp}
\usepackage{stfloats}
\usepackage{url}
\usepackage{verbatim}
\usepackage{graphicx}
\usepackage{cite}
\usepackage{booktabs}
\usepackage{graphicx}
\usepackage{pifont}

\begin{document}

\title{MUSE: Multi-Task Uncertainty-Aware Self-Supervised Encoder for
  mmWave ISAC}

\author{
    \IEEEauthorblockN{
        Nabeel Nisar Bhat,  Jalees Nehvi,
        Rafael Berkvens, 
        Jeroen Famaey
    } \\
}

\maketitle

\acrodef{XR}{Extended reality}
\acrodef{mmWave}{millimeter-wave}
\acrodef{mmGAN}{millimeter-wave GAN}
\acrodef{ISAC}{Integrated sensing and communication}
\acrodef{TL}{Transfer Learning}
\acrodef{DL}{Deep Learning}
\acrodef{DNN}{Deep Neural Network}
\acrodef{ML}{Machine Learning}
\acrodef{GAN}{Generative Adversarial Network}
\acrodef{SSL}{Self-Supervised Learning}
\acrodef{ECE}{Expected Calibration Error}
\acrodef{CNN}{Convolutional Neural Network}
\acrodef{OFDM}{Orthogonal Frequency
Division Multiplexing }
\acrodef{RFSoC}{Radio Frequency System-on-Chip}
\acrodef{FFT}{Fast Fourier Transform}
\acrodef{IFFT}{Inverse Fast Fourier Transform}
\acrodef{CP}{Cyclic prefix}
\acrodef{HCI}{Human Computer Interaction}
\acrodef{pp}{percentage points}
\acrodef{BN}{batch normalization}
\acrodef{BCE}{binary cross entropy}
\acrodef{Tx}{transmitter}
\acrodef{Rx}{receiver}
\acrodef{PPBP}{power per beam pair}
\acrodef{HW}{hardware}
\acrodef{COTS}{commercial-off-the-shelf}
\acrodef{CIR}{channel impulse response}
\acrodef{CSI}{channel state information}
\acrodef{CFO}{carrier frequency offset}

\begin{abstract}
\ac{ISAC} allows a wireless system to use the same waveform for communication and environment sensing.  At mmWave frequency bands, \ac{ISAC} enables diverse human sensing tasks, such as localization, gesture recognition, and pose estimation, with high precision. However, each task requires training a dedicated model, typically from scratch on a task-specific dataset, with no mechanism to flag unreliable predictions. This hurts the scalability, transferability, and reliability of mmWave ISAC models. There is a lack of a unified approach that learns transferable features from diverse mmWave sensing tasks while providing uncertainty estimates for reliable deployment. In this work, we propose MUSE (Multi-task Uncertainty-aware
Self-supervised Encoder), a unified framework for mmWave human sensing. MUSE leverages a self-supervised transformer model with per-dataset tokenization and is trained via masked autoencoding across five datasets. During fine-tuning, MUSE  uses evidential classification heads that enable uncertainty quantification. MUSE replaces N separate encoders with a single shared transformer encoder, an N-times backbone parameter reduction, while achieving a 9.7 percentage points improvement in classification performance compared to the task-specific model, showing that learning jointly across tasks improves per-task generalization. Across four classification tasks, MUSE produces reliable uncertainty estimates that enable selective prediction, improving accuracy by abstaining on the least confident 20\% of predictions. We further improve the reliability of the model’s confidence estimates using temperature scaling and conformal prediction. Finally, MUSE's uncertainty signal remains reliable even under cross-user deployment, assigning higher uncertainty to wrong predictions than to correct ones 78\% of the time, and recovering 9.2 percentage points of accuracy through selective prediction, with no additional retraining on the unseen user.

\end{abstract}

\begin{IEEEkeywords}
millimeter-wave, integrated sensing and communication, 5G, human sensing, gesture recognition.
\end{IEEEkeywords}
\let\thefootnote\relax\footnotetext{
Nabeel Nisar Bhat, Rafael Berkvens, and Jeroen Famaey are with the IDLab research group at the University of Antwerp - imec, Antwerp, Belgium (e-mail: firstname.lastname@uantwerpen.be). Jalees Nehvi is with the Technical University of Darmstadt (e-mail: firstname.lastname@tu-darmstadt.de). Nabeel Bhat is funded by the Fund for Scientific Research Flanders (FWO) under grant agreement number 1SH5X24N. This work was partially supported by the Hexa-X-II project, funded by the Smart Networks and Services Joint Undertaking (SNS JU) under the European Union’s Horizon Europe research and innovation programme (Grant Agreement No. 101095759), the Research Foundation - Flanders (FWO) project WaveVR (Grant No. G034322N). }
\section{Introduction}
Millimeter-wave frequency bands (24-100 GHz) \cite{de2021convergent,wymeersch2025cross} offer high bandwidth that can support data rates of several gigabits per second to enable advanced 5G and 6G communication services. At the same time, the high range and spatial resolution associated with mmWave signals can be exploited for high-precision environmental sensing. These characteristics make mmWave \acf{ISAC} a key enabler for emerging applications such as \ac{XR} and autonomous driving. In these applications, mmWave ISAC can provide advanced sensing functionalities, including device-free gesture recognition, pose estimation, environment mapping of indoor and outdoor environments, and obstacle detection 

Recent works have shown significant progress in such applications, particularly for human sensing \cite{bhat2023gesture,bhat2025mmgan,pegoraro2025disc,ren2021hand}. However, each task is typically solved in \emph{isolation} using a dedicated deep learning model trained from \textit{scratch} on a single task with no mechanism to learn shared representations across related tasks \cite{pegoraro2025disc,yang2023sensefi,bhat2023gesture,9894724,gao2022towards,dai2025meta,liu2025efficient}. This isolated-task paradigm creates a number of challenges. First, every task requires collecting and annotating a dataset and recruiting participants, which becomes particularly expensive and time-consuming. Further, there is a lack of research hardware platforms at mmWave frequency bands that enable access to signal features as \ac{CSI} \cite{pegoraro2025disc}. As a result, most of the existing mmWave datasets have a small amount of labeled data, compared to vision or text-based applications, where the dataset collection is comparatively cheaper. Because of smaller datasets, \emph{task-specific models} learn task-specific representations, resulting in poor generalization.
Further, most of the works on ISAC do not indicate the model's \emph{reliability score} i.e., a confidence score reflecting how likely a prediction is to be correct. However, deep learning models are often poorly calibrated \cite{guo2017calibration}, outputting overconfident predictions even on unseen or out-of-distribution (OOD) inputs. In critical applications such as autonomous driving, there should be a mechanism to flag unreliable predictions to prevent erroneous decisions.
Moreover, in ISAC sensing systems, deploying N independent models for N tasks adds to the computational, storage, energy, and maintenance costs. Each model is typically trained from scratch and requires its own dedicated training pipeline, hyperparameter tuning, and retraining on new tasks, resulting in significant engineering and deployment overhead that limits scalability.

Although different mmWave sensing tasks, such as gesture recognition and pose estimation, differ in objectives, they operate on a common physical phenomenon, i.e., mmWave signals reflecting from a human body. This shared signal structure suggests that common representations could be learned across tasks rather than training separate task-specific models. Self-supervised learning (SSL) \cite{krishnan2022self}, such as masked autoencoding (MAE) \cite{he2022masked}, has shown that unlabeled data can be used to derive powerful representations by masking portions of the input and forcing the model to reconstruct missing portions. This is done without requiring any actual labels. Then, these representations are used for diverse visual or language tasks. Can the same approach be applied to ISAC data? The major challenge with ISAC data is the \emph{heterogeneity} in the input data, such as in shape, signal features (\ac{CSI}, beam SNR, \ac{PPBP}, and micro-doppler), sampling frequencies, and underlying hardware imperfections. Further, these inputs may be gathered from diverse sensing geometries. This heterogeneity is significantly greater in wireless signals compared to vision SSL, where images have a standard input resolution. Consequently, applying SSL to ISAC data is significantly more challenging, since SSL methods often rely on consistent input structures and assumptions about data regularity to learn meaningful shared representations.

To address the above-mentioned limitations, we propose MUSE (Multi-task uncertainty-aware self-supervised encoder), a framework for ISAC-based human sensing. MUSE leverages MAE for SSL pretraining across five \emph{real-world} heterogeneous datasets, enabling the learning of transferable representations from unlabeled ISAC data. We evaluate the proposed framework on multiple heterogeneous ISAC datasets covering localization, gesture recognition, pose estimation, and person identification tasks. MUSE relies on a two-phase training strategy (cf., Figure \ref{fig:myfigure}). During SSL pretraining, MUSE uses masked reconstruction as a learning objective across all five datasets simultaneously. Instead of random masking of input patches, which may omit or underemphasize regions that carry critical motion and reflection information, we propose a novel \emph{saliency-based masking} approach: we mask the most informative patches based on signal variability and energy. This encourages the encoder to learn the most informative regions in the signal from incomplete observations, capturing the underlying physics of the mmWave signals. These learned representations are subsequently used for \emph{uncertainty-aware} prediction during fine-tuning.

During fine-tuning, we append \emph{evidential} classification heads \cite{sensoy2018evidential} to the encoder, which outputs a Dirichlet distribution instead of standard softmax scores, providing uncertainty estimation without the computational overhead of ensemble models \cite{lakshminarayanan2017simple} or MC dropout \cite{gal2016dropout}. Further, we apply a post-hoc \emph{calibration pipeline} to improve prediction reliability. Specifically, temperature scaling \cite{guo2017calibration} is used to correct overconfident model outputs and conformal prediction \cite{angelopoulos2023conformal} to provide statistical guarantees. Both techniques improve reliability without any additional training cost.

The contributions of our work are summarized as follows:
\begin{itemize}
    \item \textbf{Multi-task SSL for heterogeneous mmWave ISAC}: We propose MUSE, the first multi-task SSL framework designed for real-world heterogeneous mmWave ISAC while providing calibrated uncertainty estimates. We introduce a unified training pipeline, with a multi-task input sampler that balances data across datasets, a shared transformer architecture for all tasks, and an automatic loss weighting mechanism that learns per-task weights without any tuning. Consequently, a single shared encoder replaces $N$ separate per-task encoders, an $ N$- times encoder parameter reduction, while the full system (including task heads) achieves an average accuracy of 88.2\% across four classification tasks, a 9.7 percentage-point improvement over individually trained transformers.
    \item \textbf{Saliency-based MAE training:} We introduce a saliency-based masking strategy for MAE pretraining that exploits the signal structure of the mmWave signals. Instead of masking patches randomly, our method identifies the most informative regions based on signal variability and transition energy. This way, the encoder is encouraged to learn physically meaningful representations. This approach leads to better uncertainty estimates in comparison to the random masking strategy across four classification datasets, resulting in higher misclassification Area Under the Receiver Operating Characteristic (AUROC), lower expected calibration error (ECE), the smallest conformal prediction sets, and the highest selective prediction accuracy.

    \item  \textbf{Post-hoc calibration:}We apply post-hoc temperature scaling to substantially reduce calibration error by 45-70\% and conformal prediction to provide statistical coverage guarantees, resulting in a more reliable, well-calibrated model.
    
    \item \textbf{Cross-user deployment:} In practical ISAC deployments, new users will appear at test time with no labeled or training data. Standard models tested on unseen users or environments lead to a significant drop in accuracy, with no warning signal. MUSE's uncertainty-aware training provides a reliable uncertainty score even on unseen users. When the model is wrong, it assigns higher uncertainty to the wrong
  prediction than to the correct one 78\% of the time, compared to 51.5\%
  (near-random) for a standard softmax baseline. This reliable uncertainty prediction enables selective prediction, resulting in improved accuracy at \emph{zero retraining
  cost} (no labeled data from the target user).
\end{itemize}

\section{Related Work}
This section discusses related work and limitations of mmWave wireless sensing, self-supervised learning, and unified wireless models. 
\subsection{mmWave wireless sensing}
mmWave \ac{ISAC} has emerged as a powerful technology for enabling device-free environmental perception, supporting various end applications such as gesture recognition, object detection, pose estimation, and gait identification. Owing to large antenna arrays and high bandwidth, mmWave frequency bands (24-100 GHz) not only support high data rates but also enable high-precision sensing.
Ren et al. \cite{ren2021hand} studied hand gesture recognition using the IEEE 802.11ad mmWave standard. The authors built a system prototype where sensing and communication can alternate in time. Five hand gestures are considered for a total of 21 people. The range-doppler image (RDI) is then fed to a CNN-based deep learning model. The model achieves 93\% accuracy in hand gesture recognition.
Li et al. \cite{9894724} use mmWave radar for domain-independent gesture recognition. The system employs Frequency Modulated Continuous Wave (FMCW) radar to obtain a Dynamic Range Angle Image (DRAI) for gesture recognition. DRAI is then processed through an augmentation and segmentation framework. Finally, a deep learning framework is used to extract features for gesture recognition. The system achieved an average accuracy of 97.92\%, 99.18\%, and 98.76\% for new users, environments, and locations, respectively.

In our previous work \cite{bhat2025mmgan}, we demonstrated the feasibility of gesture recognition using communication-centric 5G mmWave \ac{OFDM} waveform. To address the challenges of poor generalization on unseen people, the authors proposed mmGAN, a GAN-based framework to improve the performance of mmWave ISAC systems under low-label settings. The results showed that mmGAN can improve the performance of gesture recognition by up to 25 percentage points while using only 0.6\% of labeled data.

Pegoraro et al. \cite{pegoraro2025disc} developed an mmWave ISAC testbed using the IEEE 802.11ay Wi-Fi standard. The authors extracted \ac{CIR} from the raw IQ samples. The dataset contained seven different people performing five different activities across two environments. Micro-doppler signatures derived from CIR served as input features for the CNN. The results showed high accuracy for activity classification on unseen subjects.

\textbf{Limitations of mmWave wireless sensing:} While the approach proposed by Li et al. \cite{9894724} was proven to generalize well across users, environments, and locations, it considered a radar-centric waveform, which is not suitable for use in a communication-centric ISAC system. Further, Ren et al. \cite{ren2021hand}, and Pegoraro et al. \cite{pegoraro2025disc}, consider a monostatic setup for sensing without physically separated Tx and Rx. Further, these studies rely on RDI and micro-doppler features, which are difficult to compute in \ac{COTS} devices, due to phase error caused by hardware imperfections \cite{tewes2021ws} such as carrier frequency offset (CFO). Therefore, in \ac{COTS} devices, the phase cannot be reliably tracked over time, making accurate Doppler information infeasible. In monostatic setups, CFO can be easily compensated because the transmitter and receiver share the same oscillator. However, in bistatic and multi-static setups, the phase across packets is unreliable, since each device has its own oscillator. Further, existing studies \cite{9894724,bhat2025mmgan,pegoraro2025disc,ren2021hand,bhat2023gesture}, along with many other sub-6 GHz Wi-Fi sensing works \cite{gao2022towards,gu2023attention,yang2023sensefi}, employ dedicated deep learning models and training pipelines typically focused on a single task such as gesture recognition, localization, or person identification. As a result, these methods exhibit limited transferability and often require full supervised retraining when tested on unseen people, environments, or different sensing tasks.

\subsection{Self-supervised learning and Unified Wireless Models } \ac{SSL} \cite{liu2021self} has shown promising results in vision and language tasks. SSL allows the model to learn without using actual labels during the pretraining stage. This is done by creating proxy or pretext tasks such as masking parts of the input and reconstructing the missing samples, applying transformations (e.g., rotation or permutation) and predicting the applied transformation, or forecasting future signal segments. This way, the model is forced to learn the underlying structure within the data rather than task-specific features. In computer vision, Masked Autoencoders (MAE) \cite{he2022masked}, Contrastive Learning (SimCLR) \cite{chen2020simple}, and DINO \cite{caron2021emerging} have shown promising results in self-supervised training. For example, in MAE \cite{he2022masked}, a large portion of the pixels of the image are masked, and the network is forced to reconstruct the masked pixels. This not only helps the model to learn the semantic structure of the input, leading to strong transferable representations for the downstream tasks, but also helps with faster training. Similarly, in text, transformer models such as BERT \cite{devlin2019bert} and GPT \cite{radford2018improving} learn generic representations either by predicting the masked tokens or the next word in the sequence. Recently, multi-modal models such as ImageBind \cite{girdhar2023imagebind} have shown that a single model can be used to learn joint embedding across different modalities such as images, text, audio, depth, thermal, and IMU data.

Recently, unified models have also gained traction in the ISAC community, thanks to SSL. For example, recent works such as WirelessGPT \cite{yang2025wirelessgpt}, Wireless Multimodal Foundation Model (WMFM) \cite{farzanullah2025wireless}, and foundation models for multi-task prediction \cite{sheng2025wireless} have shown promising results in LoS/NLoS classification, channel estimation, and activity recognition. 

\textbf{Limitations of unified wireless models:} 
Existing unified models \cite{yang2025wirelessgpt,sheng2025wireless,farzanullah2025wireless} suffer several limitations that MUSE aims to address. First, these models rely significantly on simulated data and do not fully capture real-world signal phenomena such as multipath and hardware imperfections. Models that learn on simulated data may not transfer well to real-world scenarios. Second, these works lack diversity in signal features, downstream sensing tasks, and sensing geometries. For example, WirelessGPT \cite{yang2025wirelessgpt}, WMFM \cite{farzanullah2025wireless}, and the work of Sheng et al. \cite{sheng2025wireless}, along with many other works such as SenseFi \cite{yang2023sensefi}, operate exclusively on CSI. Further, these offer limited evaluation as far as human-centric tasks (gesture recognition, pose estimation, or localization) are concerned. Finally, these models do not provide uncertainty estimation, which limits their reliability and deployment in safety-critical or real-world settings.

\textbf{Bridging the gap:}
To address the gap, we propose MUSE, a unified framework for heterogeneous mmWave ISAC sensing tasks. MUSE is evaluated on datasets with \emph{diverse signal features} such as \ac{CSI}, \ac{PPBP}, and beam SNR, \emph{sampling rates}, \emph{input dimensions}, \emph{sensing geometries}, \emph{environments}, \emph{hardware platforms}, and \emph{downstream tasks}. These characteristics pose a significant \emph{challenge} in developing a universal mmWave ISAC model that can learn task-agnostic transferable representations. We show that MUSE can successfully serve as a universal backbone for diverse mmWave sensing tasks. Further, MUSE enables uncertainty-aware predictions and improves the reliability of mmWave ISAC models, enabling safer deployments in real-world scenarios. To the best of our knowledge, this is the first \emph{unified wireless} model study evaluated on heterogeneous real-world mmWave sensing datasets that learns generalized representations across multiple human-centric sensing tasks, bridging the gap between simulation-based models and practical ISAC deployments. Table \ref{tab:comparison} summarizes the comparison of MUSE against state-of-the-art wireless sensing-related works. Calibration indicates whether a method's predicted confidence scores match its empirical accuracy. Reliable OOD indicates whether the model's uncertainty signal remains informative for flagging misclassifications under a cross-user setting.
\section{Problem formulation}
This section describes the problem statement and system components.
 \begin{table*}[t]                            
  \centering
  \caption{Comparison with related work.          
Note that  WirelessGPT~\cite{yang2025wirelessgpt} uses six datasets, four of which are simulated.}   
  \label{tab:comparison}                          
  \footnotesize                                   
  \setlength{\tabcolsep}{4pt}                      
  \renewcommand{\arraystretch}{1.1}
  \begin{tabular}{lcccccccc}
  \toprule
  Method & Tasks & Real & Communication & Uncertainty & Calibration & Cross- & Reliable & Signal \\
         &       & Data & Waveform &  Quantification   &        & user   & Out Of Distribution      & diversity \\                             
  \midrule                                        
  Ren et al.~\cite{ren2021hand}                   
    & 1 & \checkmark & $\times$ & $\times$ & $\times$ & $\times$ & $\times$ & $\times$ \\  
  Li et al.~\cite{9894724}                        
    & 1 & \checkmark & $\times$ & $\times$ & $\times$ & \checkmark & $\times$ & $\times$ \\                                            
  mmGAN~\cite{bhat2025mmgan}                      
    & 1 & \checkmark & \checkmark & $\times$ & $\times$ & \checkmark & $\times$ & $\times$ \\                                            
  Pegoraro et al.~\cite{pegoraro2025disc}
    & 1 & \checkmark & \checkmark & $\times$ & $\times$ & \checkmark & $\times$ & $\times$ \\                                            
  WirelessGPT~\cite{yang2025wirelessgpt}
    & 2 & \checkmark & \checkmark & $\times$ & $\times$ & $\times$ & $\times$ & $\times$ \\  
  WMFM~\cite{farzanullah2025wireless}
    & 2 & $\times$ & \checkmark & $\times$ & $\times$ & $\times$ & $\times$ & $\times$ \\                                   
  Sheng et al.~\cite{sheng2025wireless}
    & 3 & $\times$ & \checkmark & $\times$ & $\times$ & $\times$ & $\times$ & $\times$ \\    
  \midrule                                          
  \textbf{MUSE (Ours)}
    & \textbf{5} & \checkmark & \checkmark & \checkmark & \checkmark & \checkmark & \checkmark & \checkmark \\                                       
  \bottomrule                                     
  \end{tabular}
  \end{table*}

\subsection{Problem Statement}
We consider $K$ mmWave ISAC sensing datasets 
\begin{equation}
\mathcal{D}_k = \{(x_i^k, y_i^k)\}_{i=1}^{N_k}, \quad k = 1, \dots, K,
\end{equation}
where $x_i^k$ denotes the $i$-th input sample and $y_i^k$ denotes its corresponding label in the $k$-th dataset. $N_k$ represents the total number of samples in the $k$-th dataset, which varies across different datasets. 

\textbf{Objective}: The goal is to learn a unified encoder $f_\theta$, parameterized by $\theta$, that maps all inputs from $K$ datasets to a common $\mathbb{R}^d$, d-dimensional representation space. 
\begin{equation}
\min_{\theta, \{\phi_k\}, \{\omega_k\}}
\sum_{k=1}^{K}
\mathcal{L}_k \Big(
g_{\phi_k}\big(f_\theta(T_{\omega_k}(x_i^k))\big),
y_i^k
\Big)
\end{equation}
where $g_{\phi_k}$ represents a task-specific head, $T_{\omega_k}$ is a dataset-specific tokenizer, $\mathcal{L}_k$ is the task loss for dataset $k$ (e.g., cross-entropy for classification or MSE for regression),
and $f_\theta$ is shared across all $K$ datasets. This is achieved via two-stage training. However, each dataset has different input shapes; as such, we first tokenize the data using dataset-specific tokenizers ($\mathcal{T}_{\omega_k}$) \cite{ferdous2024spt}.

\subsection{System Components}
    \begin{enumerate}
    \item $\mathcal{T}_{\omega_k}$ is a per-dataset patch tokenizer, parameterized by $\omega_k$ (a Conv2d weight matrix + bias + LayerNorm parameters) that processes input data from variable shapes and converts them into a sequence of $L_k$ patch tokens, each represented as a $d$-dimensional embedding.
    \item $f_\theta : \mathbb{R}^{L_k \times d} \rightarrow \mathbb{R}^{L_k \times d}$
      is the unified transformer-based encoder shared across all $K$ datasets and parameterized by $\theta$. It processes a variable-length token sequence and produces meaningful representations.
    \item   $g_{\phi_k}: \mathbb{R}^d \rightarrow \mathcal{Y}_k$ represents a per-task head parameterized by $\phi_k$ that maps encoder representations to task-specific output space, i.e., classification or regression logits. This is only used in Phase 2.
    \end{enumerate}

\section{Methodology}
In this section, we describe patch tokenization for heterogeneous ISAC data, a unified transformer model, a multi-task training strategy, and two phases of MUSE training.
\subsection{Per-dataset patch tokenization}
\begin{figure*}[t]
    \centering
    \includegraphics[width=15cm,trim = 0cm 0cm 0cm 0cm, clip]{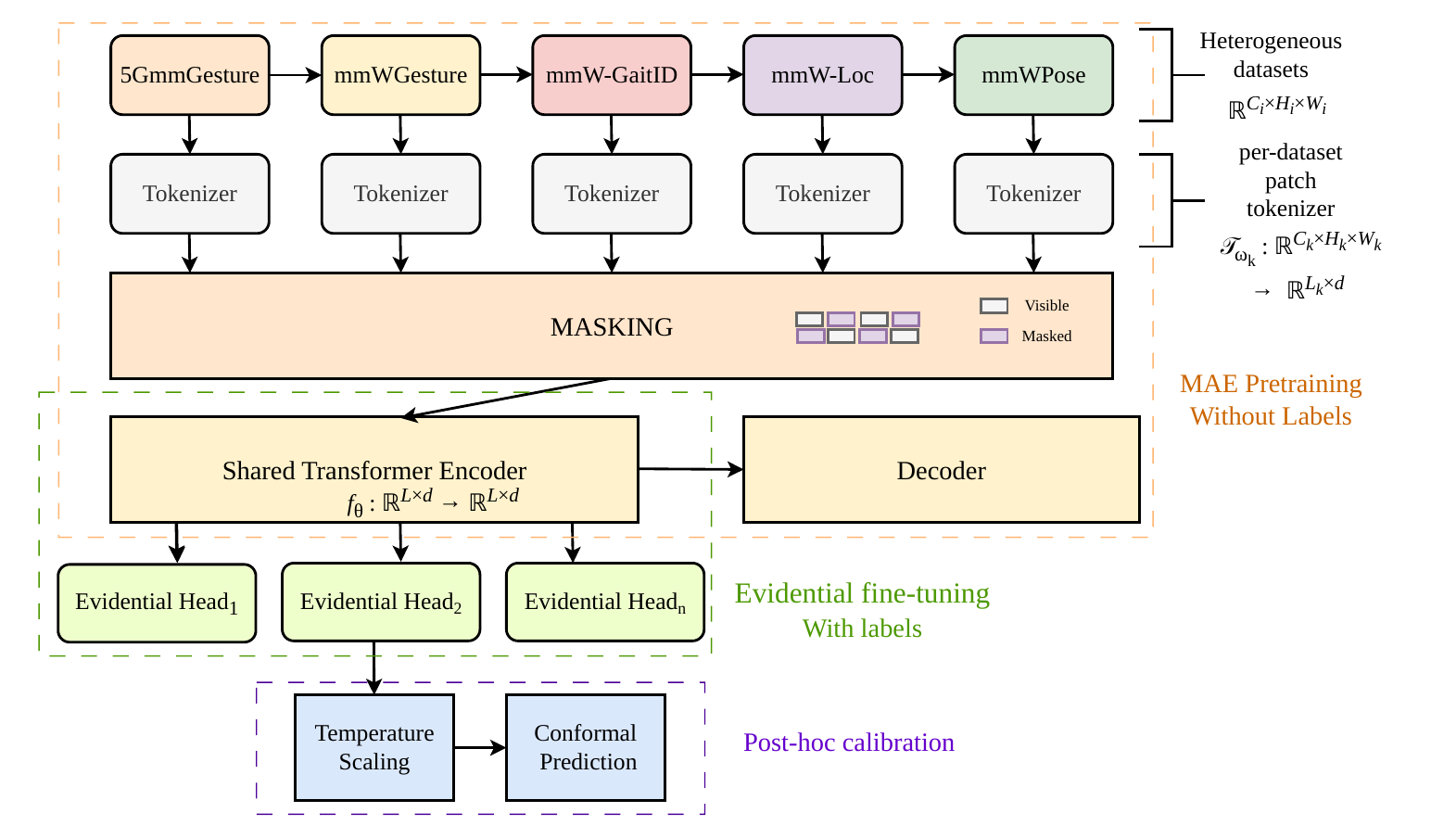}
    \caption{Overall pipeline of MUSE SSL training: Different datasets go through a per-dataset patch tokenizer. This is followed by masking and phase 1 MAE training. In phase 2, uncertainty-aware training with evidential classification heads and post hoc calibration improves the model's reliability for practical ISAC deployments.}
    \label{fig:myfigure}
\end{figure*}

Figure \ref{fig:myfigure} shows the overall pipeline of the MUSE framework. Each dataset has its own tokenizer, a convolutional layer with kernel size equal to the stride, that maps the input tensor into a $d$-dimensional embedding. Given an input tensor of shape $C_k \times H_k \times W_k$,
  the tokenizer $\mathcal{T}_{\omega_k}$ applies a single
  \texttt{Conv2d} layer followed by \texttt{LayerNorm}:

  \begin{equation}
  \begin{aligned}
  \mathcal{T}_{\omega_k}(\mathbf{x})
  &= \mathrm{LayerNorm}\Bigl(
     \mathrm{Conv2d}\bigl(\mathbf{x};\,\mathbf{W}_{\mathrm{tok}}^{(k)}\bigr)
     \Bigr), \\
  \mathbf{W}_{\mathrm{tok}}^{(k)}
  &\in \mathbb{R}^{d \times C_k \times p_k^H \times p_k^W},
  \qquad \text{stride} = (p_k^H,\, p_k^W), \\
  \mathcal{T}_{\omega_k}(\mathbf{x})
  &\in \mathbb{R}^{L_k \times d}
  \end{aligned}
  \end{equation}
Here, $\mathbf{x}$ represents the input tensor, 
$\mathbf{W}_{\mathrm{tok}}^{(k)}$ represents the learnable weights of the \texttt{Conv2d} layer, $p_k^H$ and $p_k^W$ represent the patch height and patch width, respectively. Stride equal to the patch height ensures that the input is divided into $L_k$ non-overlapping patches \cite{ferdous2024spt}.

\begin{equation}
L_k = \left\lfloor \frac{H_k}{p_k^H} \right\rfloor
      \times
      \left\lfloor \frac{W_k}{p_k^W} \right\rfloor
\end{equation}
Here, $L_k$ is the number of resulting tokens or sequence length.

\begin{table}[t]
\centering
\caption{Per-dataset patching configuration}
\label{tab:patches}
\begin{tabular}{lccc}
\toprule
Dataset & $(C_k,H_k,W_k)$ & Patch $(p_k^H,p_k^W)$ & Tokens $L_k$ \\
\midrule
5GmmGesture & $(20,50,56)$  & $(10,14)$ & 20 \\
mmWPose     & $(1,30,50)$   & $(6,10)$  & 25 \\
mmW-Loc     & $(1,60,20)$   & $(10,4)$  & 30 \\
mmW-GaitID  & $(1,60,20)$   & $(10,4)$  & 30 \\
mmWGesture  & $(1,36,500)$  & $(6,50)$  & 60 \\
\bottomrule
\end{tabular}
\end{table}
Table \ref{tab:patches} shows the patching operation on five considered datasets, with the patch size and the resulting tokens. After tokenization, all datasets produce a $ d=256$-dimensional embedding of different sequence lengths (tokens). Therefore, all \emph{input heterogeneity} is handled by $\mathcal{T}_{\omega_k}$. \texttt{LayerNorm} normalizes embeddings for each token. These embeddings are then fed to a unified transformer-based encoder, which learns task-agnostic features shared across all datasets.

\subsection{Unified Transformer Encoder}
MUSE uses a transformer encoder \cite{vaswani2017attention} with $N_L=6$ layers, and $N_H=8$ attention heads, model dimension $d=256$ and feedforward dimension $d_{\mathrm{ff}} = 1024$. These values are chosen and validated by an ablation study; increasing $d$ beyond 256 and reducing $N_L$ hurts the performance (cf., Section \ref{sec:ablation}).

  Given a tokenized sequence $\mathbf{E}^{(k)} = \mathcal{T}_{\omega_k}(\mathbf{x})
  \in \mathbb{R}^{L_k \times d}$ from dataset $k$, the encoder input is:
  \begin{equation}
    \mathbf{H}^{(0)}
    = \bigl[\,\mathbf{e}_{\mathrm{cls}};\;
             \mathbf{E}^{(k)} \,\bigr]
      \;+\; \mathbf{P}_{1:L_k+1}
      \;+\; \mathbf{m}_k \cdot \mathbf{1}^\top
    \label{eq:encoder_input}
  \end{equation}
  where $\mathbf{e}_{\mathrm{cls}} \in \mathbb{R}^{d}$ is a learnable
  class token that is appended to the input sequence, $\mathbf{P} \in \mathbb{R}^{(L_{\max}+1) \times d}$ is a learnable
  positional embedding table ($L_{\max} = 64$), sliced to length
  $L_k + 1$ and $\mathbf{m}_k \in \mathbb{R}^d$ is a \emph{learnable
  modality embedding} for dataset $k$, which is added to the input tokens. Positional embedding makes sure that every token knows its position in the original sequence, so that the model can retain spatial-temporal features from the patches.
  
Each transformer layer $\ell = 1, \ldots, N_L$ has two main blocks: Multi-head self-attention (MHSA) and Feedforward network (FFN) \cite{vaswani2017attention}. These blocks use \texttt{LayerNorm} (LN) and residual connections. With stochastic depth \cite{huang2016deep}, the output of each block can be skipped with some probability. However, residual connections ensure that some signal goes into the next block. The output of the MHSA block at layer $\ell$ is:
\begin{equation}
\mathbf{H}^{(\ell)}_{\mathrm{attn}}
= \mathbf{H}^{(\ell-1)}
  + \mathrm{DropPath}_\ell\Bigl(
      \mathrm{MHSA}\bigl(
        \mathrm{LN}(\mathbf{H}^{(\ell-1)})
      \bigr)
    \Bigr)
\label{eq:mhsa_layer}
\end{equation}

The FFN block computes the output as:
\begin{equation}
\mathbf{H}^{(\ell)}
= \mathbf{H}^{(\ell)}_{\mathrm{attn}}
  + \mathrm{DropPath}_\ell\Bigl(
      \mathrm{FFN}\bigl(
        \mathrm{LN}(\mathbf{H}^{(\ell)}_{\mathrm{attn}})
      \bigr)
    \Bigr)
\label{eq:ffn_layer}
\end{equation}
where $\mathrm{DropPath}_\ell$ applies stochastic depth with drop probability linearly increasing from $0$ to $0.1$ across layers. This allows the model to train a smaller architecture by probabilistically skipping some layers and hence acts as a form of \emph{regularization}. The output of the encoder after $N_L$ layers has the same shape as the input: $ \mathbf{H}^{(N_L)},\mathbf{H}^{(0)} \in \mathbb{R}^{L_k \times d}$.
\subsection{Multi-task training strategy} Due to heterogeneous input dimensions, standard joint batching is not feasible. Instead, we adopt \emph{sequential interleaved training}. To sample training examples from different datasets, we use a \emph{Multi-task Sampler} with temperature scaling. At each training step, dataset $k$ is sampled with probability $p_k$:
\begin{equation}
    p_k = \frac{N_k^{\,\tau}}{\sum_{j=1}^{K} N_j^{\,\tau}},
    \qquad \tau = 0.5
    \label{eq:sampling_prob}
  \end{equation}
where $\tau$ controls the trade-off between proportional sampling (where the largest dataset dominates the training loop) and uniform sampling ($\tau=0$, each dataset is equally likely) \cite{NEURIPS2019_c04c19c2}. We set $\tau=0.5$, square-root weighting, ensuring that the encoder has sufficient exposure to all datasets. A mini-batch is drawn from the training set $\mathcal{D}_k^{\mathrm{train}}$, and one complete training step (forward and backward pass) is performed through $\mathcal{T}_{\omega_k}$, $f_\theta$, and
  $g_{\phi_k}$. This is different from standard multi-task learning, where a single batch contains samples from all tasks and the overall loss is the sum of all losses in one step.  The $f_\theta$ processes all datasets in turn and receives gradient updates over an epoch in proportion to the sampling probabilities of each dataset. This applies to both Phase 1 and Phase 2. 
\subsection{Phase 1: Self-Supervised pretraining}
\paragraph{MAE training}
In this case, the goal is to reconstruct the input data using masked autoencodings, i.e., from incomplete observations of the input. 3 components participate in MAE training: tokenizer, encoder, and decoder. Given a tokenized sequence $\mathbf{E}^{(k)} \in \mathbb{R}^{L_k \times d}$ for dataset $k$, we split the sequence into a set $\mathcal{V}$ of visible tokens and a set $\mathcal{M}$ of masked tokens, where $|\mathcal{M}| = \lfloor \rho \cdot L_k \rfloor$ where $\rho = 0.5$ (50\% masking ratio). A 50\% masking ratio encourages the encoder to capture the global spatial-temporal structure of the signal. Aggressive masking ($>$75\%) is sub-optimal, since it will result in a small number of visible tokens due to small $L_k$ (20 to 60). Consequently, $\rho = 0.5$ also speeds up training. 

\textbf{Encoder pass:} After input passes through  $\mathcal{T}_{\omega_k}$, $f_\theta$ is exposed only to tokens in $\mathcal{V}$ from $K$ datasets during training. 
\begin{equation}
    \mathbf{Z}_{\mathcal{V}}
    = f_\theta\!\bigl(
        \mathbf{E}^{(k)}_{\mathcal{V}}
      \bigr)
    \;\in\; \mathbb{R}^{|\mathcal{V}| \times d}
    \label{eq:mae_encode}
  \end{equation}
where $\mathbf{Z}_{\mathcal{V}} $ represents the output of the encoder, and $\mathbf{E}^{(k)}_{\mathcal{V}} \in \mathbb{R}^{|\mathcal{V}| \times d}$ selects only the visible tokens from $\mathbf{E}^{(k)}$. Note that $e_{cls}$ is included in Eq.~\ref{eq:encoder_input} for completeness; during Phase 1 pretraining, it is excluded from the input (and thus not updated), and is only activated in Phase 2 fine-tuning. This is because MAE reconstruction does not require a global summary.

\textbf{Decoder:} The decoder is kept lightweight so that the reconstruction task becomes difficult rather than trivial. This encourages $f_\theta$ to learn meaningful representations from visible tokens. The decoder is a transformer with ($N_L^{\mathrm{dec}} = 2$ layers,
  $d^{\mathrm{dec}} = 128$, $N_H^{\mathrm{dec}} = 4$ heads). It takes the
  encoded visible tokens and learnable mask tokens
  $\mathbf{e}_{\mathrm{mask}} \in \mathbb{R}^d$, unshuffles them to
  the original order via a permutation index $\pi_{\mathrm{restore}}$,
  and reconstructs all tokens:
\begin{align}
    \tilde{\mathbf{E}}_j &=
    \begin{cases}
      \mathbf{Z}_{\mathcal{V},j}' & \text{if } j \in \mathcal{V}, \\
      \mathbf{e}_{\mathrm{mask}} + \mathbf{P}_j^{\mathrm{dec}} & \text{if } j \in \mathcal{M},
    \end{cases}
    \label{eq:unshuffle} \\[6pt]
    \hat{\mathbf{R}} &= f_{\psi}^{\mathrm{dec}}\!\bigl(
      \tilde{\mathbf{E}}
    \bigr) \;\in\; \mathbb{R}^{L_k \times d^{\mathrm{dec}}}
    \label{eq:decoder_output}
  \end{align}
 where $\mathbf{Z}_{\mathcal{V},j}'$ denotes encoder visible embeddings projected via a linear layer to the decoder dimensions $d^{dec}$, and $\mathbf{P}^{\mathrm{dec}}$ are decoder positional embeddings. $\tilde{\mathbf{E}}_j$ represents the input to decoder $f_\psi$. $\hat{\mathbf{R}}$ represents the reconstructed token sequence.
  
  A dataset-specific linear layer maps $\hat{\mathbf{R}}$ back to the input space:
\begin{equation}
\hat{\mathbf{x}}_j^{(k)} = \mathbf{W}_{\mathrm{recon}}^{(k)} \, \hat{\mathbf{R}}_j + \mathbf{b}_{\mathrm{recon}}^{(k)}, \quad 
\mathbf{W}_{\mathrm{recon}}^{(k)} \in \mathbb{R}^{(C_k \cdot p_k^H \cdot p_k^W) \times d^{\mathrm{dec}}}
\label{eq:recon_head}
\end{equation}
where $b_{recon}^{(k)}$ is a dataset-specific bias vector, and $\hat{x}_j^{(k)}$ is the reconstructed input for the $k$-th dataset and $j$-th patch

\textbf{MAE loss:}
  The loss is computed only on masked positions rather than the entire input space. This forces the encoder to understand the context and learn generalizable representations rather than simply memorizing the input.
  \begin{equation}
    \mathcal{L}_{\mathrm{MAE}}^{(k)}
    = \frac{1}{|\mathcal{M}|}
      \sum_{j \in \mathcal{M}}
      \bigl\| \hat{\mathbf{x}}_j^{(k)} - \mathbf{x}_j^{(k)} \bigr\|_2^2
    \label{eq:mae_loss}
  \end{equation}
  where $\mathbf{x}_j^{(k)} \in \mathbb{R}^{C_k \cdot p_k^H \cdot p_k^W}$
  is the flattened input content of the $j$-th patch.
\paragraph{Masking strategy -- Saliency-based masking} We introduce a saliency-based masking strategy for MAE pretraining. This strategy masks the most informative patches in the input. 
Given the $j$-th patch $\mathbf{X}_j^{(k)} \in \mathbb{R}^{C_k \times p_k^H \times p_k^W}$, we first compute the variance along $W$ and then take the average along $C$ and $H$.
\begin{align}
v_j &= \frac{1}{C_k \cdot p_k^H}
       \sum_{c=1}^{C_k} \sum_{h=1}^{p_k^H}
       \mathrm{Var}_{w=1}^{p_k^W}\!\bigl(X_{j,c,h,w}^{(k)}\bigr)\\
t_j &= \frac{1}{C_k \cdot p_k^H}
       \sum_{c=1}^{C_k} \sum_{h=1}^{p_k^H}
       \sum_{w=1}^{p_k^W - 1}
       \bigl( X_{j,c,h,w+1}^{(k)} - X_{j,c,h,w}^{(k)} \bigr)^2\\
s_j &= \frac{1}{2}\!\left(
      \frac{v_j - \min_l v_l}{\max_l v_l - \min_l v_l + \epsilon}
      + \frac{t_j - \min_l t_l}{\max_l t_l - \min_l t_l + \epsilon}
    \right)
\end{align}

where $v_j$ represents the signal variability, $t_j$ represents transition energy, and $s_j$ is the overall saliency score for the $j$-th patch. $\epsilon = 10^{-8}$ for numerical stability, and the min/max are computed per sample across all $L_k$ patches. Then, the most informative patches are masked with probabilistic Gumbel top-$k$ sampling. This adds \emph{stochasticity} to the sampling of masked patches in proportion to the saliency scores. Gumbel noise ensures that occasionally, even patches with lower saliency get selected, preventing the model from always seeing the same masking pattern, which could lead to overfitting. 

The masked patch set is defined as:
\begin{align}
\mathcal{M} &= 
\mathrm{top\text{-}k}\Biggl(
  \Bigl\{ s_j + G_j \Bigr\}_{j=1}^{L_k},\;
  k = \Bigl\lfloor \rho \cdot L_k \Bigr\rfloor
\Biggr)
\label{eq:gumbel_topk}
\end{align}
where $G_j \sim \mathrm{Gumbel}(0,1)$ represents a random sample drawn from standard Gumbel distribution.

We also compare saliency-based masking with random and curriculum-based masking. In random masking, we randomly mask $50\%$ of the input tokens.
\begin{equation}
    P(j \in \mathcal{M}) = \rho, \quad \forall j \in \{1, \ldots, L_k\}
    \label{eq:random_mask}
  \end{equation}
In curriculum-based masking, we use a schedule to gradually transition from random masking to saliency-based masking. We draw inspiration from curriculum learning\cite{bengio2009curriculum}, where the network is first exposed to easier examples, followed by difficult ones.
\begin{equation}
\resizebox{\columnwidth}{!}{$
\begin{aligned}
\text{Masking at epoch } e =
\begin{cases}
\text{Random} & e < 0.3E \quad (\text{warmup}), \\[4pt]

\begin{cases}
\text{saliency} & \text{w.p. } \frac{e - 0.3E}{0.4E} \\
\text{Random} & \text{otherwise}
\end{cases}
& 0.3E \le e < 0.7E \quad (\text{blend}), \\[10pt]

\begin{cases}
\text{Inverse saliency} & \text{w.p. } 0.3 \\
\text{saliency} & \text{otherwise}
\end{cases}
& e \ge 0.7E \quad (\text{full}).
\end{cases}
\end{aligned}
$}
\end{equation}
where $E$ represents the total number of epochs.
\subsection{Phase 2: Uncertainty-aware supervised fine-tuning}

\label{sec:FT}
In this case, we unfreeze $\theta$ and $w_k$ and instead jointly train them along with $\phi_k$. $f_\theta$ has prelearned contextual mmWave representations during MAE training, and fine-tuning adapts the structured representation space to downstream tasks for improved performance. 
After tokenization, we prepend a learnable classification token $\texttt{cls}$, denoted by $e_{\mathrm{cls}} \in \mathbb{R}^d$, to the token sequence before passing it to the encoder (cf., Eq. \ref{eq:encoder_input}). Then, the shared transformer processes all $L_k+1$ tokens through its 6 layers. Finally, the output at the last layer is defined as:
\begin{equation}
\mathbf{H}^{(N_L)} = f_{\theta}\!\left(\mathbf{H}^{(0)}\right)
\in \mathbb{R}^{(L_k+1)\times d}
\end{equation}
The \texttt{cls} token at position 0 in the sequence contains the contextual information of all other tokens via self-attention and therefore is extracted as a final representation  $z^{(k)}$, and fed to task-specific heads $g_{\phi_k}$.
\begin{equation}
\mathbf{z}^{(k)} = \mathbf{H}^{(N_L)}_{[0]}
\in \mathbb{R}^{d}
\label{eq:cls_extract_ft}
\end{equation}
To make the system uncertainty-aware, we use evidential classification heads instead of softmax heads. The evidential heads output parameters of a Dirichlet distribution. For a classification task $k$ with $C_k$ classes, the evidential heads output 
 concentration parameters 
$\boldsymbol{\alpha} = (\alpha_1, \dots, \alpha_{C_k})$, where:
\[
\alpha_c = \mathrm{softplus}(l_c) + 1
\]
where $l_c$ represents the raw logit scores and $\alpha_c$ represents the amount of evidence for a class $k$.

The class probabilities and uncertainty are computed as: 
\begin{align}
\hat{p}_c &= \frac{\alpha_c}{S},
\quad S = \sum_{c=1}^{C_k} \alpha_c \label{eq:dirichlet_prob} \\
u &= \frac{C_k}{S}\label{eq:dirichlet_uncertainty}
\end{align}
where $S$ signifies total evidence, $\hat{p}_c$ is the probability for each class, and $u$ represents epistemic uncertainty. Compared to MC dropout, $u$ is computed in a single forward pass.

To train such a model to output accurate predictions and uncertainty estimates, we optimize the following objective:
\begin{equation}
\mathcal{L}_{\text{evid}} 
= \mathcal{L}_{\text{CE}}(\boldsymbol{\alpha}, y) 
+ \lambda(e) \, \mathrm{KL}\!\left[ 
\mathrm{Dir}(\tilde{\boldsymbol{\alpha}}) 
\,\|\, \mathrm{Dir}(\mathbf{1}) 
\right]
\end{equation}
where
  $\tilde{\boldsymbol{\alpha}} = \mathbf{y} + (1 - \mathbf{y}) \odot \boldsymbol{\alpha}$
  removes evidence for the \emph{correct} class before computing KL, so the regulariser only penalises evidence assigned to
  \emph{wrong} classes.
  $\lambda(e)$ is annealed linearly from $0$ to $\lambda_{\max} = 0.1$
  over the first $100$ epochs, allowing the model to learn \emph{standard discriminative features} before applying KL regularization.
$\mathcal{L}_{\text{CE}}$ is the standard cross-entropy loss. The KL term pushes wrong-class evidence toward 1 (maximum uncertainty), while keeping the evidence for the correct class unchanged, which is instead handled by $\mathcal{L}_{\text{CE}}$.

For the regression task (mmWPose), a standard linear regression head with Mean Squared Error (MSE) loss is used, as evidential uncertainty estimation for regression requires a distinct probabilistic formulation beyond the scope of this work \cite{amini2020deep}.

\textbf{Multi-task loss weighting:} Due to heterogeneity in datasets, number of classes, and end tasks, classification and regression, the loss scales vary significantly for each task. Since the same shared encoder receives updates in an interleaved fashion from each dataset (task), tasks with high loss may dominate encoder updates, suppressing gradient updates of smaller-scale losses, irrespective of the sampling probability. To tackle this automatically, we use the Kendall weighting strategy \cite{kendall2018multi}. Each task is assigned a single learnable scalar $\sigma_k$ initialized to $\log \sigma_k = 0$. $\sigma_k$ represents the task-specific noise level that scales the loss of the respective task (dataset):
 \begin{equation}
\mathcal{L}_{\mathrm{total}}
    = \sum_{k=1}^{K}
      \left(
        \frac{1}{2\sigma_k^2} \, \mathcal{L}_{\mathrm{raw}}^{(k)}
        + \log \sigma_k
      \right)
    \label{eq:kendall_loss}
  \end{equation}
  where $\mathcal{L}_{\mathrm{raw}}^{(k)}$ is either $\mathcal{L}_{\mathrm{evid}}^{(k)}$ (for classification tasks) or
  $\mathcal{L}_{\mathrm{MSE}}^{(k)}$ (for regression task).
$\sigma_k$  is learned automatically during training. Tasks with high loss learn larger $\sigma_k$, effectively lowering their gradient contributions, while tasks with small-scale loss learn lower $\sigma_k$, increasing their gradient contributions. The rationale is that if the task is hard, predictions are noisy, and gradient updates are weakened. On the other hand, if the task is easy, predictions are less noisy, and stronger gradient updates are allowed. Thus, each task contributes comparable gradient updates to the shared encoder on average during training, without requiring manual loss normalization.

\subsection{Post-hoc calibration}
\label{sec:posthoc}
After fine-tuning, we apply a zero-cost (without any training overhead) post-hoc calibration pipeline that improves the reliability of the model's predictions.
\paragraph{Temperature scaling} Evidential classification heads are overconfident, meaning that the confidence score is higher than the actual accuracy. For example, for all the test samples, the model assigns a probability of 0.9 to its top prediction, but it may be correct only 70\% of the time. This calibration gap can be measured using Expected Calibration Error (ECE) \cite{roelofs2022mitigating}, which quantifies the gap between the model's confidence and actual accuracy, across all confidence bins. To tackle this calibration error, we apply post-hoc temperature scaling per dataset.
\paragraph{Conformal Coverage}

We apply post-hoc split conformal prediction \cite{oliveira2024split} to provide statistical guarantees on individual predictions. Rather than outputting a single prediction, conformal prediction constructs a prediction set, with guarantees to contain a true class with probability at least $1 - \varepsilon$, where $\varepsilon$ represents the error rate. When the model is confident, the prediction set is a singleton. However, in case of uncertainty, the prediction set may be two or more classes, \emph{signalling} failure of the model.

\section{Experimental setup}
This section describes the five considered datasets, experimental configuration, baselines, and MUSE variant models used for comparison.
\subsection{Datasets}
\begin{table}[t]
\centering
\caption{Overview of real-world mmWave ISAC human sensing datasets.}
\label{tab:datasets}
\footnotesize
\renewcommand{\arraystretch}{1.05}
\setlength{\tabcolsep}{2pt}

\begin{tabular}{l c c c r r r}
\toprule
\textbf{Dataset} & \textbf{Sig.} & \textbf{Shape} & \textbf{Task} &
\textbf{Lbls.} & \textbf{Samples} & \textbf{Tokens} \\
\midrule
5GmmGesture & PPBP     & $20\times50\times56$  & Cls. & 8    & 36,344 & 20 \\
mmWPose     & CSI Amp. & $1\times30\times50$   & Reg. & 3,750 & 4,035  & 25 \\
mmW-Loc     & CSI Amp. & $1\times60\times20$   & Cls. & 20   & 2,177  & 30 \\
mmW-GaitID  & CSI Amp. & $1\times60\times20$   & Cls. & 20   & 1,318  & 30 \\
mmWGesture  & Beam SNR & $1\times36\times500$  & Cls. & 10   & 865    & 60 \\
\midrule
\textbf{Total} & -- & -- & -- & -- & \textbf{44,739} & -- \\
\bottomrule
\end{tabular}
\end{table}

Table \ref{tab:datasets} describes the overview of five datasets we consider, with signal features used such as \ac{PPBP}, CSI amplitude (CSI Amp), input shape of the signal, downstream task (Class. representing classification, and Regr. representing regression), number of output labels, samples, and tokens.
\begin{enumerate}
    \item \textbf{5GmmGesture:} is a dataset collected using a 5G mmWave NR testbed in a bi-static configuration. The Tx transmits a 5G OFDM waveform at 60 GHz frequency. The input is a tensor of shape $C \times H \times W$ representing \ac{PPBP}, with $C=20$ representing time samples, $H=50$ representing Tx beams, and $W=56$ representing Rx beams. 8 subjects perform 8 gestures. PPBP has a sampling frequency of 154 Hz.
    \item \textbf{mmWPose:} is collected using IEEE 802.11ad \ac{COTS} testbed in a bi-static configuration. The input represents CSI amplitude with 30 antenna elements and 50 time samples, sampled at 22 Hz. Kinect is used to obtain ground truth for skeletal-based pose estimation. The regression target has 3,750 dimensions (25 body joints × 3 coordinates × 50 time steps).
    \item \textbf{mmW-Loc:} also collected using the IEEE 802.11ad testbed; however, two pairs of Tx-Rx are used in X-configuration. Input is CSI amplitude, from 60 antennas (representing a concatenation of two 30 antennas) and 20 time steps. The end-task is indoor localization across a $4\times5$ spatial grid, a total of 20 positions. 20 different people participated in this experiment. The sampling frequency of CSI is 13 Hz.
    \item \textbf{mmW-GaitID}: collected using same setup as mmW-Loc. The end task is person identification by observing walking patterns (gait). A total of 20 people participated in this data collection.
    \item \textbf{mmWGesture}: collected using IEEE 802.11ad \ac{COTS} testbed in bi-static configuration. Beam SNR is extracted as a signal feature from the sector sweep procedure. The end task is gesture recognition across 10 gestures. The datasets consist of three people and two environments, and one rotation-based experiment. The sampling frequency of the beam SNR is 10 Hz. 
\end{enumerate}
  For more details about the datasets and experimental settings, please refer to our previous work \cite{11511681}.  All of the five datasets involve real-world experiments that capture diversities of signal features, sensing geometries, signal type, and input shapes, making it particularly difficult for a shared encoder to learn transferable representations across different tasks and modalities.

\subsection{Configuration}
All experiments were run on NVIDIA Tesla V100-SXM3-32GB GPU, with PyTorch 2.7.1 and CUDA 11.8. For
reproducibility, we set \texttt{torch.backends.cudnn.deterministic = True}
and \texttt{torch.backends.cudnn.benchmark = False}.
For classification tasks, we split the dataset into 80/10/10 train, validation, and test splits, respectively, using stratified splits preserving class ratios. Batch size is set to 32.  Moreover, for each experiment, we use 3 different seeds (42, 123, and 456). After training, the model is evaluated on the test set based on the best validation accuracy. All the models, including baselines, MUSE variations such as Linear Probe, MAE training, and full fine-tuning, are trained for 200 epochs. For the baselines, standard cross-entropy loss is used for classification tasks and MSE for regression.

\subsection{Baselines}
\textbf{Individual transformer models:}
In this case, we use the exact same architecture, with a per-dataset tokenizer, $\mathcal{T}_{\omega_k}$, the encoder $f_\theta$, and the same task head $g_{\phi_k}$ as the foundation fine-tuning case. The key difference here is that every dataset gets its own $f_\theta$ that is trained from scratch on the respective labels only. This results in 5 separate models with 5 encoders, a total of 23.8M parameters ($5\times 4.76$M) for the encoder alone, whereas MUSE relies on a single encoder with 4.76M parameters. Other settings such as architecture, hyperparameters (AdamW, lr=$5 \times 10^{-4}$,
weight decay 0.01, epochs=200), data splits, and seeds are kept identical to the full fine-tuning setting.

\textbf{ResNet-18:} ResNet-18 \cite{he2016deep} represents the standard CNN baseline with the first \texttt{Conv} layer adapted to each dataset's input shape. The fully connected (\texttt{fc}) layer is replaced by a two-layer output head: Linear(512, 256) $\to$ ReLU $\to$ Dropout(0.1) $\to$ Linear(256, $C_k$). This is also trained from scratch per dataset with the same optimizer settings as MUSE
fine-tuning (AdamW, lr=$5 \times 10^{-4}$, weight decay 0.01, 200 epochs).

\textbf{LSTM}: 2-layer Bi-directional LSTM  \cite{chadha2020bidirectional} with $H$ dimension as sequence length and $C\times W$ as feature dimension. A dropout of 0.3 is applied between the two layers. Final forward and backward hidden states are concatenated, resulting in a 512-dimensional representation vector, which is fed to a 2-layer output head: Linear(512, 256) $\to$ ReLU $\to$ Dropout(0.1) $\to$ Linear(256, $C_k$). We train a bi-directional LSTM from scratch per dataset using the same optimizer settings.

\subsection{MUSE Variants}
We consider several variants of MUSE, in addition to full fine-tuning:
\paragraph{Linear probing} After Phase 1, we freeze the $f_\theta$ and $w_k$, and a single linear layer is trained on top of the encoder representations. To obtain representations from the encoder, we mean-pool the $L$ patch-token outputs of the encoder: $\mathbf{z}^{(k)} = \frac{1}{L}\sum_{i=1}^{L}
f_\theta\!\left(\mathcal{T}_{\omega_k}(\mathbf{x})\right)_i
  \in \mathbb{R}^{d}.$
where $d=256$. Accordingly, the linear classifier takes 256-dimensional inputs, while the output dimension is dataset-dependent. This represents the strictest test of representation quality.

\paragraph{Frozen encoder + MLP heads} Here again, $\theta$ and $w_k$ are kept frozen, and a 2-layer MLP head is trained on top of mean-pooled representations similar to linear probing. In particular, the head is composed of: \texttt{LayerNorm} → \texttt{Linear} → \texttt{GELU} → \texttt{Dropout} → \texttt{Linear}. The same is true for the regression task with the mmWPose dataset; only the last layer is adapted depending on the output. This introduces approximately $70K$ trainable parameters per classification task, $280K$ for 4 classification tasks.

\paragraph{Frozen encoder + Prompt tuning for classification tasks} In this case, we investigate parameter-efficient fine-tuning, prompt tuning \cite{lester2021power} for ISAC, in particular for the classification tasks. For each dataset, we introduce $P=4$ learnable prompt tokens, prepended to the tokenized patches before feeding them to the encoder. This value is chosen to add minimal overhead ($4\times d = 1,024$ parameters).  Smaller values ($P=1,2$) provide insufficient conditioning signal for diverse sensing tasks \cite{jia2022visual}. Notably, $\theta$ and $w_k$ are frozen here as well.
\begin{equation}
E^{(k)}_{\text{prompt}} =
\left[
e^{k}_1, \ldots, e^{k}_P,
T_{\omega_k}(x)_1, \ldots, T_{\omega_k}(x)_L
\right]
\in \mathbb{R}^{(P+L)\times d}
\end{equation}
Although encoder weights are frozen, the self-attention mechanism allows the patch tokens to \emph{attend} to learnable task-specific prompts at each layer in the encoder. These prompts act as conditions and guide the encoder to produce \emph{task-specific} representations. At the end of the output of the encoder, the output associated with prompt tokens is discarded, and the remaining tokens are mean-pooled to obtain a task-conditioned representation. This is then passed to a shared decoder, a 2-layer MLP decoder, and a task-specific linear projection. Compared to the MLP-based approach, this significantly reduces the number of trainable parameters.
\section{Results and Discussion}
This section presents and discusses experimental results. We compare MUSE against several baselines in terms of accuracy and computational cost, followed by an analysis of masking strategies. Finally, we present results related to prompt tuning and hyperparameter sensitivity.
\begin{table*}[t]
\centering
\caption{Performance comparison across five mmWave ISAC tasks (mean over three random seeds).
Classification tasks are evaluated using accuracy ($\uparrow$), while the pose estimation task is evaluated using mean squared error (MSE, $\downarrow$).
Datasets: 5GmmGesture (5G), mmW-GaitID (Gait), mmW-Loc (Loc), mmWGesture (mWG), and mmWPose (Pose).
Avg denotes the mean accuracy over the four classification tasks.
\#Param denotes the total number of trainable parameters across all five tasks. Std-H represents the standard softmax head. Evid-H represents evidential head.}
\label{tab:main_results}
\setlength{\tabcolsep}{4pt}
\renewcommand{\arraystretch}{1.25}

\begin{tabular}{l l c c c c c c r}
\toprule
\multirow{2}{*}{Config.}
  & \multirow{2}{*}{Method}
  & \multicolumn{4}{c}{Classification Accuracy $\uparrow$}
  & Regression
  & \multirow{2}{*}{Avg $\uparrow$}
  & \multirow{2}{*}{\#Param (total)} \\
\cmidrule(lr){3-6} \cmidrule(lr){7-7}
  & & 5G & Gait & Loc & mWG & Pose MSE $\downarrow$ & & \\
\midrule

Repr.\ Quality
  & Linear Probe & 0.783 & 0.799 & 0.499 & 0.391 & 0.0167 & 0.618 & 979K \\
\midrule

\multirow{3}{*}{\shortstack[l]{Supervised\\Baselines\\(separate)}}
  & Transformer & 0.973 & 0.819 & 0.705 & 0.644 & 0.0025 & 0.785 & 27M \\
  & ResNet-18   & \textbf{0.980} & 0.814 & 0.609 & \textbf{0.847} & 0.0017 & 0.813 & 57M \\
  & Bi-LSTM     & 0.973 & 0.743 & 0.560 & 0.295 & 0.0053 & 0.643 & 16M \\
\midrule

\multirow{2}{*}{\shortstack[l]{MUSE\\(Frozen Encoder)}}
  & Frozen+MLP    & 0.907 & \underline{0.896} & 0.734 & 0.755 & 0.0041 & 0.823 & 2.34M \\
  & Frozen+Prompt & 0.913 & 0.883 & 0.720 & 0.709 & -- & 0.807 & 151K \\
\midrule

\multirow{4}{*}{\shortstack[l]{MUSE\\(Full FT)}}
  & Rand-MAE + Std-H  & \underline{0.979} & \textbf{0.906} & 0.805 & \underline{0.839} & \underline{0.0015} & \textbf{0.882} & 7.9M \\
  & Sal.-MAE + Evid-H & 0.976 & 0.870 & \underline{0.811} & 0.797 & \textbf{0.0014} & \underline{0.864} & 7.9M \\
  & Curr-MAE + Evid-H & 0.975 & 0.873 & \textbf{0.813} & 0.743 & 0.0016 & 0.851 & 7.9M \\
  & Rand-MAE + Evid-H & 0.977 & 0.878 & \textbf{0.813} & 0.759 & 0.0016 & 0.857 & 7.9M \\
\bottomrule
\end{tabular}
\end{table*}

\subsection{Accuracy and Computational Cost}
Table \ref{tab:main_results} shows the performance of MUSE across five heterogeneous tasks. Full FT refers to full fine-tuning, where shared encoders, tokenizers, and classification heads are jointly trained. Rand-MAE refers to random masking, Curr-MAE: Curriculum masking, and Sal. MAE: saliency masking (referring to MAE pretraining strategy). Repr.Quality refers to the representation quality evaluated using a linear probe \cite{he2022masked,chen2020simple}. Std-H refers to standard softmax classification heads. Evid-H refers to evidential heads with uncertainty quantification. Bold represents the best per column, and underlined values indicate the second best per column.

\textbf{Linear probing:} Linear probing represents the strictest evaluation of representation quality. As can be seen, on the 5GmmGesture and mmW-GaitID, the linear probe achieves a decent accuracy of 78\% and 80\%, respectively. However, on the other three tasks, the performance is poor, suggesting that the representations are not linearly separable and therefore requiring a non-linear head to improve the performance. For this setting, the number of trainable parameters is 979K, the second lowest in Table \ref{tab:main_results}.

\textbf{Baselines: }Among the baselines, there is no single baseline that performs well on all tasks. For example, on the 5GmmGesture dataset, which has the most training examples, all baselines perform decently well. However, all models struggle on the mmW-GaitID and mmW-Loc datasets, which have comparatively fewer training examples. Both the transformer and the Bi-LSTM struggle with the mmWGesture dataset. Overall, ResNet-18 seems to perform better on average across all tasks. However, this configuration requires training 5 separate models, leading to a total parameter count of 57M, 27M, and 16M for ResNet-18, Transformer, and Bi-LSTM, respectively. 

\textbf{Frozen encoder settings:} In the frozen encoder configuration, we see that the MLP-based classification head performs decently well on the 5GmmGesture dataset and the mmWGaitID, even better than the best baseline in the latter case. While for the regression task, the performance is significantly worse. Nevertheless, on average, this method beats the best baseline ResNet-18 on classification tasks, requiring only 2.34M parameters (24 times less than ResNet-18's 57M). Further, using prompt tuning (prompt), we achieve nearly similar average accuracy on the classification tasks as the MLP-based classification head while requiring 1.8 times fewer parameters (151K vs 280K for MLP \emph{excluding pose task}). 
\begin{table}[t]
\centering
\caption{Standard deviation ($\sigma$) of the performance metrics computed over three random seeds.}
\label{tab:std_results}
\footnotesize
\setlength{\tabcolsep}{2.5pt}
\renewcommand{\arraystretch}{1.1}

\begin{tabular}{lccccc}
\toprule
\multirow{2}{*}{Method}
& \multicolumn{4}{c}{Accuracy $\sigma$}
& \multirow{2}{*}{MSE $\sigma$} \\
\cmidrule(lr){2-5}
& 5G & Gait & Loc & mWG & \\
\midrule

Rand-MAE + Std-H  & 0.003 & 0.016 & 0.025 & 0.019 & 0.0003 \\
Sal.-MAE + Evid-H & 0.001 & 0.015 & 0.015 & 0.057 & 0.0003 \\
Curr-MAE + Evid-H & 0.005 & 0.009 & 0.025 & 0.013 & 0.0003 \\
Rand-MAE + Evid-H & 0.002 & 0.013 & 0.005 & 0.035 & 0.0003 \\
\midrule

Frozen+MLP        & 0.008 & 0.016 & 0.008 & 0.020 & 0.0002 \\
Frozen+Prompt     & 0.011 & 0.019 & 0.026 & 0.052 & -- \\
\midrule

Transformer       & 0.002 & 0.025 & 0.030 & 0.073 & 0.0003 \\
ResNet-18         & 0.003 & 0.031 & 0.043 & 0.052 & 0.0002 \\
Bi-LSTM           & 0.001 & 0.031 & 0.005 & 0.037 & 0.0005 \\
\midrule

Linear Probe      & 0.004 & 0.023 & 0.015 & 0.064 & 0.0005 \\

\bottomrule
\end{tabular}
\end{table}
Both results suggest that frozen encoder representations generalize effectively across different downstream tasks, despite a smaller computational cost compared to the baselines.

\textbf{Full fine-tuning:} 
In the full fine-tuning case, we see that MUSE performs consistently well on all tasks compared to the supervised baselines and frozen encoder settings. Across all models, MUSE full FT variants achieve the best performance on three out of five datasets. Compared to the best baseline, we achieve an 8.7 percentage points improvement on the mmW-GaitID dataset, a 10 percentage points improvement on the mmW-Loc dataset, and also a lower MSE on mmWPose.  The benefit of multi-task SSL is more pronounced for smaller datasets. For mmWGesture and 5GmmGesture datasets, the performance is comparable with the best baseline. Across different configurations of MUSE full fine-tuning, variants with evidential classification heads slightly trade performance for the ability to quantify uncertainty. On average across 4 classification tasks, this model achieves a 5 percentage points improvement compared to the best baseline ResNet-18. Furthermore, compared to the transformer baseline, MUSE full fine-tuning provides a 10 percentage points improvement with identical architecture, tokenizers, encoder, and classification head, same hyperparameter settings. The only difference is that MUSE's encoder is shared across all five tasks     
rather than trained independently per dataset. Specifically, one encoder (4.76M) replaces five separate encoders (23.78M), a 5 times backbone parameter count reduction, bringing the full system to 7.9M parameters, 3.4$\times$ fewer than five individually trained transformers (27M). This confirms that the performance gains are attributable to MUSE's multi-task pretraining and shared encoder design. Note that standard deviations ($\sigma$) (cf., Table~\ref{tab:std_results}) are small across most datasets, with MUSE variants generally showing low variance. The elevated variance on mmWGesture is observed for most methods, reflecting dataset-level complexity rather than model instability.
\begin{table}[t]
\centering
\caption{Comparison of masking strategies in terms of uncertainty quantification and model reliability: Evidential fine-tuning.}
\label{tab:masking_comparison}
\setlength{\tabcolsep}{6pt}
\renewcommand{\arraystretch}{1.15}
\begin{tabular}{lccc}
\toprule
Metric & Saliency & Curriculum & Random \\
\midrule
Accuracy $\uparrow$ & \textbf{0.864} & 0.851 & 0.857 \\
Misclassification AUROC $\uparrow$ & \textbf{0.700} & 0.685 & 0.677 \\
ECE $\downarrow$ & \textbf{0.122} & 0.131 & 0.128 \\
Conformal Set Size $\downarrow$ & \textbf{1.241} & 1.289 & 1.273 \\
Selective @90\% $\uparrow$ & \textbf{0.893} & 0.878 & 0.881 \\
Selective @80\% $\uparrow$ & \textbf{0.912} & 0.897 & 0.901 \\
Selective @50\% $\uparrow$ & \textbf{0.929} & 0.908 & 0.909 \\
\bottomrule
\end{tabular}
\end{table}
\subsection{Masking and uncertainty analysis}
We evaluate uncertainty quality using several metrics:
\begin{itemize}
   \item \textbf{Misclassification AUROC}: the area under the ROC curve for detecting wrong predictions using uncertainty as 
the score (0.5 = random detector, 1.0 = perfect).
   \item \textbf{Expected Calibration Error (ECE)}: the average gap between predicted confidence and empirical accuracy across 
confidence bins (lower is better).
\item \textbf{Conformal Set Size}: the average number of classes in the conformal prediction set at 90\% coverage (1.0 = 
always singleton, most certain) \cite{oliveira2024split}.
\item \textbf{Selective Accuracy @K\%} \cite{geifman2017selective}: accuracy when the model answers only the $K\%$ most 
confident predictions.
\end{itemize}
Table \ref{tab:masking_comparison} shows the performance of three masking strategies averaged across 3 seeds and four classification tasks. All three masking strategies, random, curriculum, and saliency-based masking, are trained identically. We can see that all metrics are better for saliency-based masking, improving misclassification AUROC by 1.5 and 2.3 percentage points, and reducing ECE by 7\% and 5\%, respectively, relative to curriculum and random masking, and yielding the smallest mean conformal set size. This suggests that saliency-based masking results in a more reliable uncertainty-aware model. 

We dig deeper into uncertainty analysis by looking at per-dataset metrics and saliency-based masking. Table \ref{tab:phys_evid_per_dataset} shows the same metrics for each dataset separately. For the mmW-GaitID and mmW-Loc, the misclassification AUROC is significantly higher, suggesting that the model knows well when it is making wrong predictions. For the 5GmmGesture dataset, it is nearly random; this is due to the fact that the model is making only a few errors at 97.6\% accuracy. ECE scores suggest that for the 5GmmGesture dataset, the model is fairly well calibrated, whereas on other datasets, the model calibration error is higher, particularly for the mmWGesture dataset, which also happens to be the hardest task. We see that the conformal set size is lowest for 5GmmGesture. Finally, selective accuracy@80\% (Sel.@80\%), i.e., removing 20\% of the most uncertain predictions, improves most on mmW-GaitID (9.2 percentage points) and mmW-Loc (8.6 percentage points), and mmWGesture (8.4 percentage points). On the other hand, smaller gains (1.3 percentage points) on the 5GmmGesture dataset are due to lower AUROC and already near-ceiling accuracy, leaving few errors to abstain from.

We hypothesize that saliency-based masking allows the encoder to learn physically meaningful representations by forcing it to reconstruct the most informative patches in the signal. To reconstruct these patches, the encoder can not rely on superficial correlation or background statistics. In other words, the model is encouraged to learn the underlying mmWave signal structure, rather than memorizing signal patterns. During inference, deviations from these physically meaningful features lead to higher uncertainty, improving the model's reliability. In contrast, random masking treats all patches equally, allowing the encoder to rely on less informative signal regions. Thus, the representations are less sensitive to deviations in mmWave signal structure, meaning that the uncertainty score rises less consistently when the model is wrong, leading to less reliable uncertainty estimation.

\subsection{Post-Hoc Calibration}  
\begin{figure*}[t]
    \centering
    \includegraphics[width=14.5cm]{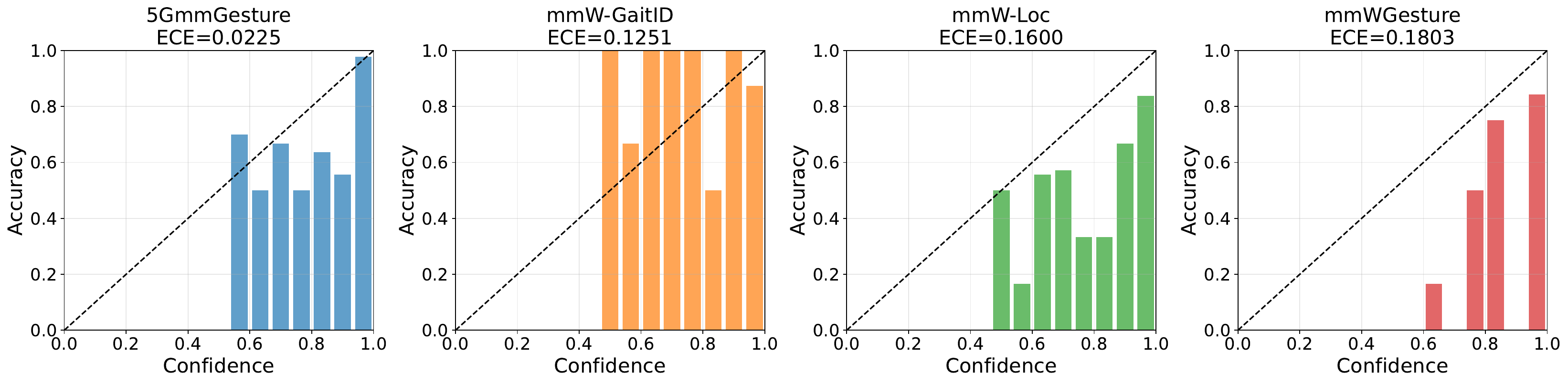}
    \caption{ECE before temperature scaling: Poor calibration of confidence scores.}
    \label{fig:ecebefore}
\end{figure*}

\begin{figure*}[t]
    \centering
    \includegraphics[width=14.5cm]{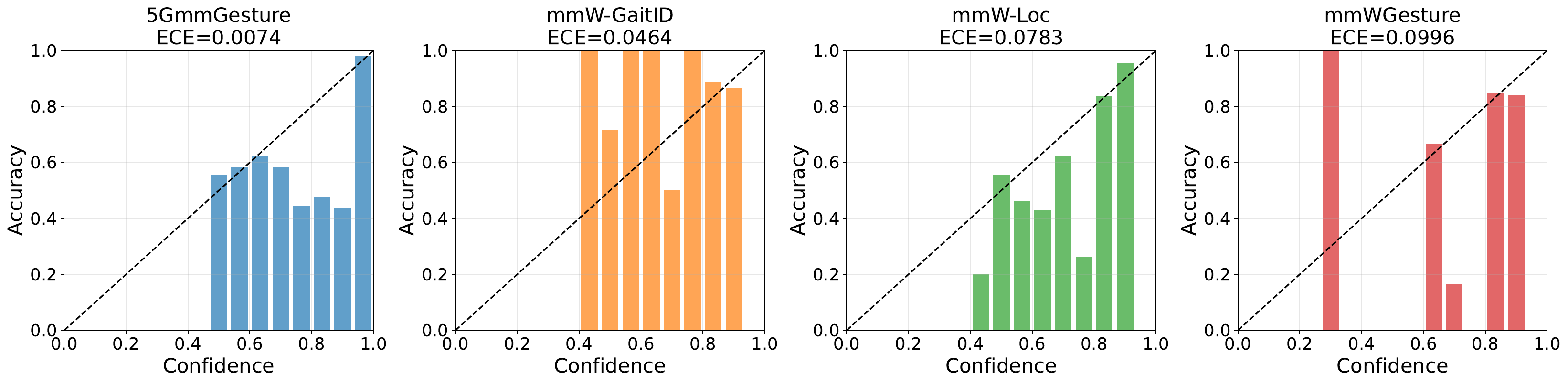}
    \caption{ECE after temperature scaling: Better calibrated model.}
    \label{fig:eceafter}
\end{figure*}

Modern neural networks lack \emph{calibration} \cite{guo2017calibration}; i.e., the predicted confidence scores are typically higher than the empirical accuracy. A well-calibrated model is one whose confidence scores match accuracy; if the neural network says it is 90\% confident, it should be correct 90\% of the time. Figure \ref{fig:ecebefore} shows the calibration error (ECE) of the MUSE's classification heads on four datasets. The y-axis represents the accuracy, and the x-axis represents model confidence $\max_c \hat{p}_c$ (cf., Eq. 22). Ideally, bars should follow the diagonal. Deviation above the diagonal indicates underconfidence, while deviation below indicates overconfidence. Table \ref{tab:ece_reduction} further shows that before calibration, MUSE's evidential classification heads are overconfident across all datasets ($T$ $>$ 1). As described in Section \ref{sec:posthoc}, we apply temperature scaling by fitting a single scalar T per dataset on the validation set. After scaling, ECE drops significantly. Figure \ref{fig:eceafter} shows the calibration error after temperature scaling. We can see that temperature scaling reduces ECE by 45-70\% across 4 datasets, improving the reliability of the model.
\begin{table}[t]
\centering
\caption{Per-dataset performance of saliency-based masking. Acc: Accuracy.}
\label{tab:phys_evid_per_dataset}
\setlength{\tabcolsep}{5pt}
\renewcommand{\arraystretch}{1.15}
\begin{tabular}{lccccc}
\toprule
Dataset & Acc $\uparrow$ & AUROC $\uparrow$ & ECE $\downarrow$ & Set Size $\downarrow$ & Sel.@80\% $\uparrow$ \\
\midrule
5G  & 0.976 & 0.553 & 0.023 & 1.103 & 0.989 \\
GaitID & 0.870 & 0.874 & 0.125 & 1.250 & 0.962 \\
Loc    & 0.811 & 0.767 & 0.160 & 1.357 & 0.897 \\
mmWG & 0.797 & 0.606 & 0.180 & 1.253 & 0.881 \\
\midrule
Mean       & 0.864 & 0.700 & 0.122 & 1.241 & 0.912 \\
\bottomrule
\end{tabular}
\end{table}
\begin{table}[t]
\centering
\caption{ECE before and after calibration, averaged over three seeds.}
\label{tab:ece_reduction}
\setlength{\tabcolsep}{5pt}
\renewcommand{\arraystretch}{1.1}
\resizebox{\columnwidth}{!}{
\begin{tabular}{lcccc}
\toprule
Dataset & ECE Before $\downarrow$ & ECE After $\downarrow$ & Reduction (\%) $\uparrow$ & Fitted $T$ \\
\midrule
5GmmGesture  & 0.023 & \textbf{0.007} & 70 & 1.90 \\
mmW-GaitID   & 0.125 & \textbf{0.046} & 63 & 1.54 \\
mmW-Loc      & 0.160 & \textbf{0.078} & 51 & 1.57 \\
mmWGesture   & 0.180 & \textbf{0.100} & 45 & 1.84 \\
\bottomrule
\end{tabular}
}
\end{table}

\begin{table}[t]
\centering
\caption{Conformal prediction performance, averaged over three seeds.}
\label{tab:conformal_results}
\setlength{\tabcolsep}{5pt}
\renewcommand{\arraystretch}{1.1}

\begin{tabular*}{\columnwidth}{l@{\extracolsep{\fill}}ccc}
\toprule
Dataset & Coverage (\%) $\uparrow$ & Mean Set Size $\downarrow$ & Singleton (\%) $\uparrow$ \\
\midrule
5GmmGesture & 98.1 & 1.10 & 90.9 \\
mmW-GaitID  & 88.5 & 1.25 & 82.2 \\
mmW-Loc     & 85.8 & 1.36 & 80.7 \\
mmWGesture  & 83.5 & 1.25 & 84.3 \\
\midrule
Avg        & 89.0 & 1.24 & 84.5 \\
\bottomrule
\end{tabular*}
\end{table}

 Table \ref{tab:conformal_results} shows the conformal prediction results. With $\alpha =0.10$, targeting 90\% coverage, MUSE achieves 89\% empirical coverage with a mean prediction set size of 1.24. Moreover, 84.5\% predictions produce a singleton (outputting one class only). Thus, in the majority of cases, conformal prediction predicts a precise result. In the remaining predictions, the model is not sure and hence outputs 2-3 possibilities, \emph{flagging} cases when the model is uncertain. This avoids committing to a wrong prediction. Temperature scaling and conformal prediction make MUSE a practically \emph{reliable framework} for ISAC systems.
\subsection{Cross-user deployment}
One of the main challenges in ISAC is performance degradation under cross-user or OOD scenarios, where the training distribution is not \emph{representative} of the test set. This occurs due to diverse body morphology, gesture execution style, and other user-specific biases. To see the benefits of MUSE's evidential classification heads on cross-user deployment, we use the 5GmmGesture dataset, consisting of 8 users. We compare it against a softmax baseline. For the cross-user experiment, we train the same architecture as MUSE full fine-tuning, but replace the evidential head with a standard softmax trained with cross-entropy loss. Uncertainty is then given by:
\begin{equation}
\label{eq:usoftmax}
u_{\text{sfmx}}(\mathbf{l}) = 1 - \max_{c} \, \mathrm{softmax}(\mathbf{l})_c
\end{equation} 
where $\mathbf{l}$ is the vector of class logits, and $c \in \{1, \ldots, C\}$ is the class index. As a first step, we set the in-domain upper bound by training and testing on all users. We get an overall average accuracy of 97.7\% across 8 users and 8 gestures.

Next, we evaluate the \emph{leave-one-user-out} setting, where the training set consists of seven users and the test set consists of the unseen (left-out) user. Table \ref{tab:loo_ood} shows the results under the leave-one-user-out protocol. We can see that the OOD accuracy (OOD Acc) drops significantly compared to the in-domain setting. Although the accuracy drops per held-out user, the uncertainty signal remains fairly strong. On average, the model has 0.78 misclassification AUROC (Miscl. AUROC), meaning that the model knows fairly well when it makes wrong predictions. Further, if the 20\% most uncertain predictions are removed (Sel@80\%), there is a 9.2 percentage points gain in accuracy on average. A higher misclassification AUROC implies a higher gain. For example, User 4 has the highest AUROC, thus a higher gain of 11.6 percentage points. This gain in accuracy is zero-cost: no additional training, no access to labeled data from unseen users. For the softmax baseline, we use Eq. \ref{eq:usoftmax} for the uncertainty score:  a low maximum softmax probability signals low confidence. From Table \ref{tab:loo_ood}, we see that softmax misclassification AUROC is 0.515 on average, suggesting that softmax confidence is no better than random guessing for identifying misclassifications in an OOD scenario.

\begin{figure}[t]
    \centering
    \includegraphics[width=8.0cm]{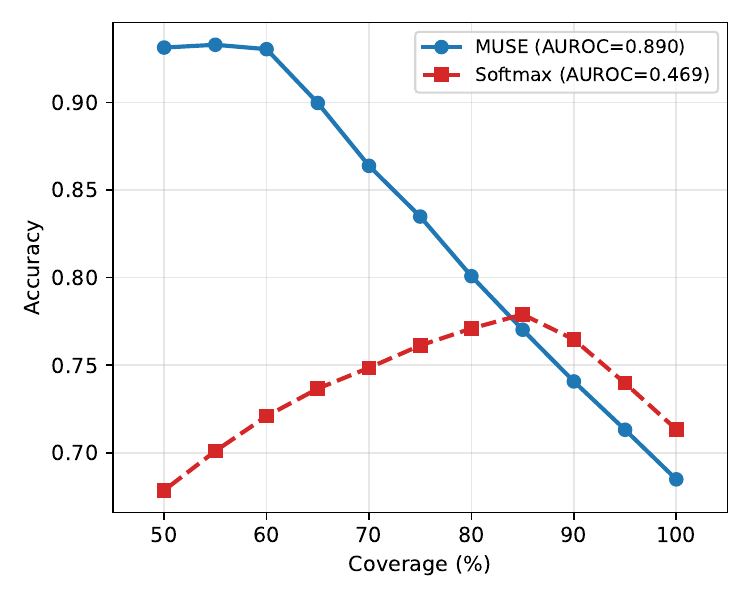}
    \caption{Accuracy vs coverage (user 4): MUSE vs Standard Softmax.}
    \label{fig:accvscoverage}
\end{figure}

\begin{figure}[t]
    \centering
    \includegraphics[width=8.0cm]{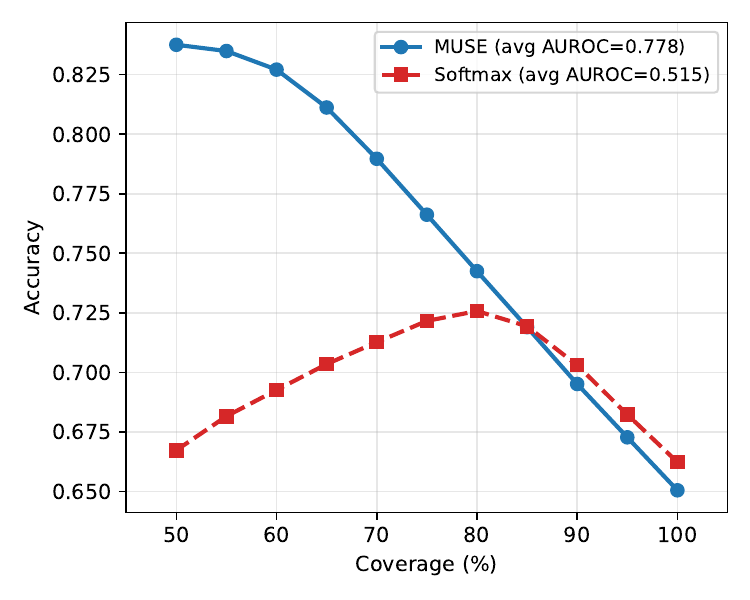}
    \caption{Accuracy vs coverage (all users): MUSE vs Standard Softmax.}\label{fig:accvscoverageallusers}
\end{figure}

Figure \ref{fig:accvscoverage} shows the accuracy as a function of coverage for held-out user 4. We see that in the case of MUSE with evidential heads,  the selective accuracy increases monotonically as the coverage drops. In contrast, softmax selective accuracy peaks at 85\% coverage and then experiences a sharp drop, suggesting that softmax confidence is not a reliable signal for identifying wrong predictions. A similar trend is observed across all users. A similar trend holds when averaged across all 8 users.  Figure~\ref{fig:accvscoverageallusers} confirms this: MUSE's accuracy again increases monotonically as coverage drops, while softmax peaks near 80\% coverage before dropping off.

\begin{figure}[t]
    \centering
    \includegraphics[width=8.0cm]{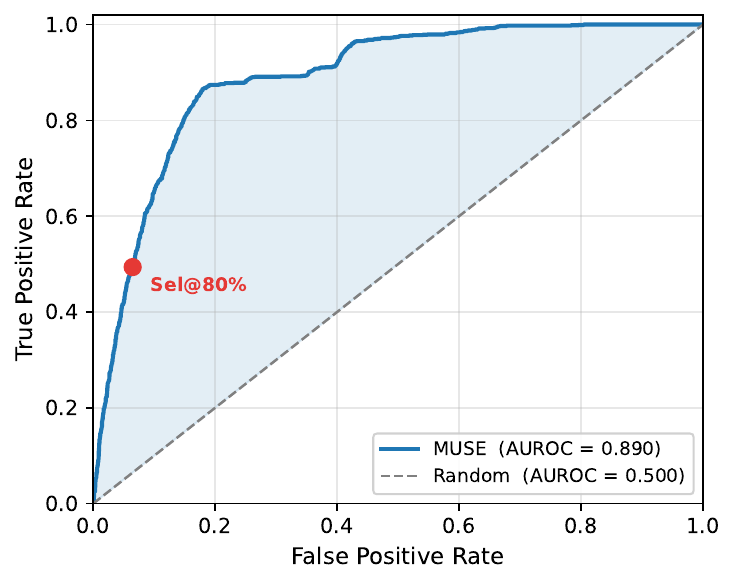}
    \caption{Misclassification detection AUROC: Held-out user 4.}
    \label{fig:roccurve}
\end{figure}

Additionally, Figure \ref{fig:roccurve} shows the misclassification detection AUROC for the same held-out user 4. The x-axis represents the fraction of correct predictions abstained from (false positive rate), while the y-axis depicts the fraction of wrong predictions correctly identified. The curve sweeps through different values of uncertainty thresholds. At each threshold, the x-axis represents the cost, and the y-axis represents the benefit. The shaded area represents the gain over random guessing. For example, at the threshold corresponding to Sel@80\%, 49\% of mistakes are correctly identified at the cost of abstaining from 6.5\% of correct predictions. AUROC remains 0.89, which means, on average, for any randomly chosen pair of one correct and one wrong prediction, there is an 89\% chance the wrong prediction has higher uncertainty than the correct one. The same results hold for other held-out users. Thus, uncertainty scores serve as a filter to \emph{selectively abstain} from the most uncertain predictions, thereby improving accuracy on unseen users at zero retraining cost.  
\begin{table}[!t]
\centering
\caption{Leave-one-user-out (OOD) evaluation averaged over three seeds. Miscl. refers to Misclassification detection.}
\label{tab:loo_ood}
\setlength{\tabcolsep}{5pt}
\begin{tabular}{lcccccc}
\toprule
& \multicolumn{1}{c}{Accuracy} 
& \multicolumn{2}{c}{Miscl.(AUROC)} 
& \multicolumn{2}{c}{Selective Prediction MUSE} \\
\cmidrule(lr){2-2} \cmidrule(lr){3-4} \cmidrule(lr){5-6}
User & OOD Acc $\uparrow$ & MUSE $\uparrow$ & Softmax $\uparrow$ & Sel@80\% $\uparrow$ & Gain@80\% \\
\midrule
1 & 0.604 & 0.750 & 0.551 & 0.684 & 8.0 \\
2 & 0.648 & 0.692 & 0.551 & 0.728 & 8.0 \\
3 & 0.782 & 0.801 & 0.500 & 0.894 & 11.2 \\
4 & 0.685 & 0.890 & 0.469 & 0.801 & 11.6 \\
5 & 0.636 & 0.796 & 0.508 & 0.717 & 8.1 \\
6 & 0.666 & 0.842 & 0.545 & 0.772 & 10.6 \\
7 & 0.636 & 0.830 & 0.465 & 0.730 & 9.4 \\
8 & 0.548 & 0.624 & 0.534 & 0.615 & 6.7 \\
\midrule
\textbf{Avg} & \textbf{0.651} & \textbf{0.778} & \textbf{0.515} & \textbf{0.743} & \textbf{9.2} \\
\bottomrule
\end{tabular}
\end{table}

\subsection{Prompt-tuning: Scaling new tasks}
For the results shown in Table \ref{tab:main_results}, we trained the shared MLP decoder, prompt tokens, and lightweight projections on all four classification tasks. Retraining the shared MLP decoder is the main cost when adding a new sensing task.  Under prompt-tuning, we train the shared decoder on the three tasks: 5GmmGesture, mmWGesture, and mmW-Loc, and leave mmW-GaitID out (Phase 1). In Phase 2, we freeze the shared decoder and add only 4 \emph{new prompt tokens} and a projection layer for mmWGait-ID. The encoder is kept frozen under both phases. This simulates a \emph{realistic deployment} scenario where a new sensing task arrives post-training and must be integrated without modifying any shared parameters.
\begin{table}[t]
\centering
\caption{Comparison of methods in terms of accuracy (mmW-GaitID) and trainable parameters.}
\label{tab:model_efficiency}
\setlength{\tabcolsep}{5pt}
\renewcommand{\arraystretch}{1.1}
\resizebox{\columnwidth}{!}{
\begin{tabular}{lcc}
\toprule
Method & Accuracy $\uparrow$ & Trainable Params \\
\midrule
MLP — 4-task joint             & \textbf{0.8855} & $\sim$280K total \\
Prompt tuning — 4-task joint       & 0.8321 & 151K total \\
Frozen decoder + new prompts & 0.7634 & \textbf{6,165 total} \\
\bottomrule
\end{tabular}
}
\end{table}
Table \ref{tab:model_efficiency} shows the results in terms of accuracy and trainable parameters. With only 6,165 trainable parameters, the model achieves 76.3\% accuracy on the mmW-GaitID, recovering 91.7\% of the jointly-trained upper bound (83.2\%) while being approximately 11 times cheaper than a per-task MLP head (6K vs.\ 70K parameters -   ($280\text{K} \div 4$)). For reference, the frozen MLP baseline achieves 88.5\% accuracy on the mmW-GaitID dataset, while requiring 70K parameters per task and 280K in total (cf., Table \ref{tab:model_efficiency}). Our result highlights that the shared MLP decoder generalizes fairly well to new sensing tasks. This is because mmW-GaitID shares a similar CSI amplitude signal format as mmW-Loc, which the decoder had already seen in Phase~1, in addition to generalized representations provided by the unified encoder. Further, the cost of adding a new sensing task is just 6K parameters, with no retraining of any shared component required, corresponding to 24 KB of memory footprint and making it lightweight enough to operate without GPU resources.

\subsection{Hyperparameter Sensitivity}
\label{sec:ablation}
\begin{table}[t]
  \centering
  \caption{Architecture sensitivity analysis (physics saliency masking, mean over 3 seeds).}
  \label{tab:ablation}
  \setlength{\tabcolsep}{4pt}
  \begin{tabular}{lcccccc}
  \toprule
  Config & 5G& Gait & Loc & mWG & Pose & Avg \\
  \midrule
  \multicolumn{7}{l}{\textit{Model dimension $d$}} \\
  $d$=256~(default) & \textbf{0.976} & 0.870 & 0.811 & \textbf{0.797} & \textbf{0.0014} & \textbf{0.864} \\
  $d$=512           & 0.977 & 0.875 & \textbf{0.825} & 0.747 & 0.0015 & 0.856 \\
  $d$=1024          & 0.963 & 0.860 & 0.699 & 0.728 & 0.0025 & 0.812 \\
  \midrule
  \multicolumn{7}{l}{\textit{Attention heads $N_H$}} \\
  $N_H$=4           & 0.969 & \textbf{0.888} & 0.796 & 0.759 & 0.0018 & 0.853 \\
  $N_H$=8~(default) & \textbf{0.976} & 0.870 & \textbf{0.811} & \textbf{0.797} & \textbf{0.0014} & \textbf{0.864} \\
  \midrule
  \multicolumn{7}{l}{\textit{Encoder depth $N_L$}} \\
  $N_L$=2           & 0.970 & 0.868 & 0.767 & 0.716 & 0.0019 & 0.830 \\
  $N_L$=3           & 0.973 & \textbf{0.885} & 0.791 & 0.785 & 0.0014 & 0.859 \\
  $N_L$=6~(default) & \textbf{0.976} & 0.870 & 0.811 & \textbf{0.797} & \textbf{0.0014} & \textbf{0.864} \\
  $N_L$=7           & 0.974 & 0.863 & \textbf{0.820} & 0.770 & 0.0014 & 0.857 \\
  \bottomrule
  \end{tabular}
  \end{table}

Table \ref{tab:ablation} shows architecture sensitivity of MUSE over $d$ dimensions, $N_H$ attention heads, and $N_L$ layers, varying one parameter at a time with all others at default ($d$=256, $N_H$=8, $N_L$=6). Increasing $d$ beyond 256 decreases the model's performance, with $d=1024$ losing 5 percentage points on average accuracy across classification tasks, and pose MSE also increases substantially with respect to the default configuration. We select $d=256$ to balance model capacity and dataset size. $N_H=4$ leads to a 1.1 percentage points loss in average accuracy. Finally, we also vary the depth by changing $N_L$. $N_L=2$ is insufficient to capture a spatial-temporal structure. Increasing depth improves performance; however, beyond $N_L=6$, the performance drops. In general, excluding the two cases ( $N_L=2$ and $d=1024$), the configurations remain within 1.2 percentage points of the default one, indicating that MUSE is relatively robust to moderate changes in architectural hyperparameters.
  
\section{Conclusion}
In this work, we proposed MUSE, a multi-task uncertainty-aware self-supervised encoder for mmWave ISAC human sensing. MUSE addresses limitations in existing ISAC models, such as task-specific training, no uncertainty awareness, poor scalability, and a lack of evaluation on diverse real-world datasets. Using 5 real-world human sensing mmWave datasets, MUSE's encoder shared among 5 tasks achieves better performance than individual baselines. On the classification tasks, MUSE improves accuracy by 9.7 percentage points compared to transformer-based individual baselines, while using 3.4 times reduced parameter count. Further, our framework enables efficient post-deployment adaptation to new sensing tasks through lightweight prompt tuning, in which only small projection heads (and prompt tokens) need to be trained for a new task, while achieving competitive performance relative to jointly trained task-specific heads. In addition, we saw that saliency-based masking combined with evidential training provided the best uncertainty metrics compared to random and curriculum-based masking. Using zero-training-cost post-hoc calibration, ECE was reduced by 45-70\%. Finally, MUSE's uncertainty signal remained reliable even under cross-user deployment, achieving a misclassification AUROC of 0.78. This was confirmed by a selective prediction test, where accuracy increased monotonically as coverage reduced, suggesting that the uncertainty score correctly identifies the wrong predictions. In contrast, softmax-based uncertainty suffered random collapse, making it an unreliable indicator of the model's failure. This makes the MUSE deployment-ready model for reliable mmWave ISAC.

\bibliographystyle{IEEEtran}
\bibliography{biblio}

\begin{thebibliography}{10}
\providecommand{\url}[1]{#1}
\csname url@samestyle\endcsname
\providecommand{\newblock}{\relax}
\providecommand{\bibinfo}[2]{#2}
\providecommand{\BIBentrySTDinterwordspacing}{\spaceskip=0pt\relax}
\providecommand{\BIBentryALTinterwordstretchfactor}{4}
\providecommand{\BIBentryALTinterwordspacing}{\spaceskip=\fontdimen2\font plus
\BIBentryALTinterwordstretchfactor\fontdimen3\font minus
  \fontdimen4\font\relax}
\providecommand{\BIBforeignlanguage}[2]{{%
\expandafter\ifx\csname l@#1\endcsname\relax
\typeout{** WARNING: IEEEtran.bst: No hyphenation pattern has been}%
\typeout{** loaded for the language `#1'. Using the pattern for}%
\typeout{** the default language instead.}%
\else
\language=\csname l@#1\endcsname
\fi
#2}}
\providecommand{\BIBdecl}{\relax}
\BIBdecl

\bibitem{de2021convergent}
C.~De~Lima, D.~Belot, R.~Berkvens, A.~Bourdoux, D.~Dardari, M.~Guillaud,
  M.~Isomursu, E.-S. Lohan, Y.~Miao, A.~N. Barreto \emph{et~al.}, ``Convergent
  communication, sensing and localization in {6G} systems: An overview of
  technologies, opportunities and challenges,'' \emph{IEEE Access}, vol.~9, pp.
  26\,902--26\,925, 2021.

\bibitem{wymeersch2025cross}
H.~Wymeersch, N.~Tervo, S.~W{\"a}nstedt, S.~Saleh, J.~Ahlendorf, O.~Akgul,
  V.~Tsekenis, S.~Barmpounakis, L.~Bai, M.~Beale \emph{et~al.}, ``Cross-layer
  integrated sensing and communication: A joint industrial and academic
  perspective,'' \emph{IEEE Open Journal of the Communications Society}, 2025.

\bibitem{bhat2023gesture}
N.~N. Bhat, R.~Berkvens, and J.~Famaey, ``Gesture recognition with {mmWave}
  {Wi-Fi} access points: Lessons learned,'' in \emph{2023 IEEE 24th
  International Symposium on a World of Wireless, Mobile and Multimedia
  Networks (WoWMoM)}.\hskip 1em plus 0.5em minus 0.4em\relax IEEE, 2023, pp.
  127--136.

\bibitem{bhat2025mmgan}
N.~N. Bhat, S.~Kumar, M.~H. Moghaddam, J.~Struye, J.~O. Lacruz, J.~Pegoraro,
  J.~Widmer, R.~Berkvens, and J.~Famaey, ``{mmGAN}: Semi-supervised {GAN} for
  improved gesture recognition in {mmWave} {ISAC} systems,'' \emph{IEEE Open
  Journal of the Communications Society}, vol.~7, pp. 95--117, 2025.

\bibitem{pegoraro2025disc}
J.~Pegoraro, P.~Saucedo, J.~O. Lacruz, M.~Rossi, and J.~Widmer, ``{DISC}: A
  dataset for integrated sensing and communications in {mmWave} systems,''
  \emph{IEEE Communications Magazine}, 2025.

\bibitem{ren2021hand}
Y.~Ren, J.~Lu, A.~Beletchi, Y.~Huang, I.~Karmanov, D.~Fontijne, C.~Patel, and
  H.~Xu, ``Hand gesture recognition using 802.11 ad {mmWave} sensor in the
  mobile device,'' in \emph{2021 IEEE wireless communications and networking
  conference workshops (WCNCW)}.\hskip 1em plus 0.5em minus 0.4em\relax IEEE,
  2021, pp. 1--6.

\bibitem{yang2023sensefi}
J.~Yang, X.~Chen, H.~Zou, C.~X. Lu, D.~Wang, S.~Sun, and L.~Xie, ``{SenseFi}: A
  library and benchmark on deep-learning-empowered {Wi-Fi} human sensing,''
  \emph{Patterns}, vol.~4, no.~3, 2023.

\bibitem{9894724}
Y.~Li, D.~Zhang, J.~Chen, J.~Wan, D.~Zhang, Y.~Hu, Q.~Sun, and Y.~Chen,
  ``Towards domain-independent and real-time gesture recognition using {mmWave}
  signal,'' \emph{IEEE Transactions on Mobile Computing}, vol.~22, no.~12, pp.
  7355--7369, 2023.

\bibitem{gao2022towards}
R.~Gao, W.~Li, Y.~Xie, E.~Yi, L.~Wang, D.~Wu, and D.~Zhang, ``Towards robust
  gesture recognition by characterizing the sensing quality of {Wi-Fi}
  signals,'' \emph{Proceedings of the ACM on Interactive, Mobile, Wearable and
  Ubiquitous Technologies}, vol.~6, no.~1, pp. 1--26, 2022.

\bibitem{dai2025meta}
P.~Dai, J.~Zhou, J.~Ma, H.~Zhang, and X.~Wu, ``Meta-transfer learning based
  cross-domain gesture recognition using wifi channel state information,''
  \emph{IEEE Transactions on Consumer Electronics}, 2025.

\bibitem{liu2025efficient}
J.~Liu, J.~Yuan, G.~Yu, and J.~Han, ``Efficient one-shot gesture recognition
  for wifi isac via aug-meta learning,'' \emph{IEEE Journal on Selected Areas
  in Communications}, 2025.

\bibitem{guo2017calibration}
C.~Guo, G.~Pleiss, Y.~Sun, and K.~Q. Weinberger, ``On calibration of modern
  neural networks,'' in \emph{International conference on machine
  learning}.\hskip 1em plus 0.5em minus 0.4em\relax PMLR, 2017, pp. 1321--1330.

\bibitem{krishnan2022self}
R.~Krishnan, P.~Rajpurkar, and E.~J. Topol, ``Self-supervised learning in
  medicine and healthcare,'' \emph{Nature Biomedical Engineering}, vol.~6,
  no.~12, pp. 1346--1352, 2022.

\bibitem{he2022masked}
K.~He, X.~Chen, S.~Xie, Y.~Li, P.~Doll{\'a}r, and R.~Girshick, ``Masked
  autoencoders are scalable vision learners,'' in \emph{Proceedings of the
  IEEE/CVF conference on computer vision and pattern recognition}, 2022, pp.
  16\,000--16\,009.

\bibitem{sensoy2018evidential}
M.~Sensoy, L.~Kaplan, and M.~Kandemir, ``Evidential deep learning to quantify
  classification uncertainty,'' \emph{Advances in neural information processing
  systems}, vol.~31, 2018.

\bibitem{lakshminarayanan2017simple}
B.~Lakshminarayanan, A.~Pritzel, and C.~Blundell, ``Simple and scalable
  predictive uncertainty estimation using deep ensembles,'' \emph{Advances in
  neural information processing systems}, vol.~30, 2017.

\bibitem{gal2016dropout}
Y.~Gal and Z.~Ghahramani, ``Dropout as a bayesian approximation: Representing
  model uncertainty in deep learning,'' in \emph{international conference on
  machine learning}.\hskip 1em plus 0.5em minus 0.4em\relax PMLR, 2016, pp.
  1050--1059.

\bibitem{angelopoulos2023conformal}
A.~N. Angelopoulos and S.~Bates, ``Conformal prediction: A gentle
  introduction,'' \emph{Foundations and Trends in Machine Learning}, vol.~16,
  no.~4, pp. 494--591, 2023.

\bibitem{tewes2021ws}
S.~Tewes and A.~Sezgin, ``{WS-WiFi}: Wired synchronization for csi extraction
  on {COTS}-{Wi-Fi}-transceivers,'' \emph{IEEE Internet of Things Journal},
  vol.~8, no.~11, pp. 9099--9108, 2021.

\bibitem{gu2023attention}
Y.~Gu, H.~Yan, X.~Zhang, Y.~Wang, J.~Huang, Y.~Ji, and F.~Ren,
  ``Attention-based gesture recognition using commodity {Wi-Fi} devices,''
  \emph{IEEE Sensors Journal}, vol.~23, no.~9, pp. 9685--9696, 2023.

\bibitem{liu2021self}
X.~Liu, F.~Zhang, Z.~Hou, L.~Mian, Z.~Wang, J.~Zhang, and J.~Tang,
  ``Self-supervised learning: Generative or contrastive,'' \emph{IEEE
  transactions on knowledge and data engineering}, vol.~35, no.~1, pp.
  857--876, 2021.

\bibitem{chen2020simple}
T.~Chen, S.~Kornblith, M.~Norouzi, and G.~Hinton, ``A simple framework for
  contrastive learning of visual representations,'' in \emph{International
  conference on machine learning}.\hskip 1em plus 0.5em minus 0.4em\relax PmLR,
  2020, pp. 1597--1607.

\bibitem{caron2021emerging}
M.~Caron, H.~Touvron, I.~Misra, H.~J{\'e}gou, J.~Mairal, P.~Bojanowski, and
  A.~Joulin, ``Emerging properties in self-supervised vision transformers,'' in
  \emph{Proceedings of the IEEE/CVF international conference on computer
  vision}, 2021, pp. 9650--9660.

\bibitem{devlin2019bert}
J.~Devlin, M.-W. Chang, K.~Lee, and K.~Toutanova, ``{BERT}: Pre-training of
  deep bidirectional transformers for language understanding,'' in
  \emph{Proceedings of the 2019 conference of the North American chapter of the
  association for computational linguistics: human language technologies,
  volume 1 (long and short papers)}, 2019, pp. 4171--4186.

\bibitem{radford2018improving}
A.~Radford, K.~Narasimhan, T.~Salimans, I.~Sutskever \emph{et~al.}, ``Improving
  language understanding by generative pre-training,'' 2018.

\bibitem{girdhar2023imagebind}
R.~Girdhar, A.~El-Nouby, Z.~Liu, M.~Singh, K.~V. Alwala, A.~Joulin, and
  I.~Misra, ``{ImageBind}: One embedding space to bind them all,'' in
  \emph{Proceedings of the IEEE/CVF conference on computer vision and pattern
  recognition}, 2023, pp. 15\,180--15\,190.

\bibitem{yang2025wirelessgpt}
T.~Yang, P.~Zhang, M.~Zheng, Y.~Shi, L.~Jing, J.~Huang, and N.~Li,
  ``{WirelessGPT}: A generative foundation model for multi-task integrated
  sensing and communication,'' \emph{IEEE Journal on Selected Areas in
  Communications}, 2025.

\bibitem{farzanullah2025wireless}
M.~Farzanullah, H.~Zhang, A.~B. Sediq, A.~Afana, and M.~Erol-Kantarci,
  ``Wireless multimodal foundation model (wmfm): Integrating vision and
  communication modalities for {6G} {ISAC} systems,'' \emph{arXiv preprint
  arXiv:2512.23897}, 2025.

\bibitem{sheng2025wireless}
Y.~Sheng, J.~Wang, X.~Zhou, L.~Liang, H.~Ye, S.~Jin, and G.~Y. Li, ``A wireless
  foundation model for multi-task prediction,'' \emph{arXiv preprint
  arXiv:2507.05938}, 2025.

\bibitem{ferdous2024spt}
G.~J. Ferdous, K.~A. Sathi, M.~A. Hossain, and M.~A.~A. Dewan, ``Spt-swin: A
  shifted patch tokenization swin transformer for image classification,''
  \emph{IEEE Access}, vol.~12, pp. 117\,617--117\,626, 2024.

\bibitem{vaswani2017attention}
A.~Vaswani, N.~Shazeer, N.~Parmar, J.~Uszkoreit, L.~Jones, A.~N. Gomez,
  {\L}.~Kaiser, and I.~Polosukhin, ``Attention is all you need,''
  \emph{Advances in neural information processing systems}, vol.~30, 2017.

\bibitem{huang2016deep}
G.~Huang, Y.~Sun, Z.~Liu, D.~Sedra, and K.~Q. Weinberger, ``Deep networks with
  stochastic depth,'' in \emph{European conference on computer vision}.\hskip
  1em plus 0.5em minus 0.4em\relax Springer, 2016, pp. 646--661.

\bibitem{NEURIPS2019_c04c19c2}
\BIBentryALTinterwordspacing
A.~CONNEAU and G.~Lample, ``Cross-lingual language model pretraining,'' in
  \emph{Advances in Neural Information Processing Systems}, H.~Wallach,
  H.~Larochelle, A.~Beygelzimer, F.~d\textquotesingle Alch\'{e}-Buc, E.~Fox,
  and R.~Garnett, Eds., vol.~32.\hskip 1em plus 0.5em minus 0.4em\relax Curran
  Associates, Inc., 2019. [Online]. Available:
  \url{https://proceedings.neurips.cc/paper_files/paper/2019/file/c04c19c2c2474dbf5f7ac4372c5b9af1-Paper.pdf}
\BIBentrySTDinterwordspacing

\bibitem{bengio2009curriculum}
Y.~Bengio, J.~Louradour, R.~Collobert, and J.~Weston, ``Curriculum learning,''
  in \emph{Proceedings of the 26th annual international conference on machine
  learning}, 2009, pp. 41--48.

\bibitem{amini2020deep}
A.~Amini, W.~Schwarting, A.~Soleimany, and D.~Rus, ``Deep evidential
  regression,'' \emph{Advances in neural information processing systems},
  vol.~33, pp. 14\,927--14\,937, 2020.

\bibitem{kendall2018multi}
A.~Kendall, Y.~Gal, and R.~Cipolla, ``Multi-task learning using uncertainty to
  weigh losses for scene geometry and semantics,'' in \emph{Proceedings of the
  IEEE conference on computer vision and pattern recognition}, 2018, pp.
  7482--7491.

\bibitem{roelofs2022mitigating}
R.~Roelofs, N.~Cain, J.~Shlens, and M.~C. Mozer, ``Mitigating bias in
  calibration error estimation,'' in \emph{International Conference on
  Artificial Intelligence and Statistics}.\hskip 1em plus 0.5em minus
  0.4em\relax PMLR, 2022, pp. 4036--4054.

\bibitem{oliveira2024split}
R.~I. Oliveira, P.~Orenstein, T.~Ramos, and J.~V. Romano, ``Split conformal
  prediction and non-exchangeable data,'' \emph{Journal of Machine Learning
  Research}, vol.~25, no. 225, pp. 1--38, 2024.

\bibitem{11511681}
N.~N. Bhat, M.~Karnaukh, S.~Vandenbroeke, W.~Lemoine, J.~Struye, S.~Kumar,
  M.~H. Moghaddam, J.~O. Lacruz, J.~Widmer, R.~Berkvens, and J.~Famaey,
  ``{mmHSense}: Multi-modal and distributed {mmWave} {ISAC} datasets for
  {Human} {Sensing},'' \emph{IEEE Access}, vol.~14, pp. 69\,961--69\,971, 2026.

\bibitem{he2016deep}
K.~He, X.~Zhang, S.~Ren, and J.~Sun, ``Deep residual learning for image
  recognition,'' in \emph{Proceedings of the IEEE conference on computer vision
  and pattern recognition}, 2016, pp. 770--778.

\bibitem{chadha2020bidirectional}
G.~S. Chadha, A.~Panambilly, A.~Schwung, and S.~X. Ding, ``Bidirectional deep
  recurrent neural networks for process fault classification,'' \emph{ISA
  transactions}, vol. 106, pp. 330--342, 2020.

\bibitem{lester2021power}
B.~Lester, R.~Al-Rfou, and N.~Constant, ``The power of scale for
  parameter-efficient prompt tuning,'' in \emph{Proceedings of the 2021
  conference on empirical methods in natural language processing}, 2021, pp.
  3045--3059.

\bibitem{jia2022visual}
M.~Jia, L.~Tang, B.-C. Chen, C.~Cardie, S.~Belongie, B.~Hariharan, and S.-N.
  Lim, ``Visual prompt tuning,'' in \emph{European conference on computer
  vision}.\hskip 1em plus 0.5em minus 0.4em\relax Springer, 2022, pp. 709--727.

\bibitem{geifman2017selective}
Y.~Geifman and R.~El-Yaniv, ``Selective classification for deep neural
  networks,'' \emph{Advances in neural information processing systems},
  vol.~30, 2017.

\end{thebibliography}

\end{document}